\documentclass{article}
\usepackage{epsfig}

\begin{document}

\long\def\comment#1{}
\newcommand{\etc}{{\em etc}}
\newcommand{\eg}{{\em e.g.}}
\newcommand{\ie}{{\em i.e.}}
\newcommand{\ea}{{\em et al.}}
\newcommand{\noi}{\noindent}

\newenvironment{itm}{\begin{itemize}\setlength{\itemsep}{-0.02in plus 0in}}{\end{itemize}}

\title{Mathematical Analysis of Multi-Agent Systems\thanks{The research reported here was supported
in part by the Defense Advanced Research Projects Agency (DARPA) under contract
number F30602-00-2-0573, in part by the National Science Foundation under Grant
No. 0074790, and by the ISI/ISD Research Fund Award.
The views and conclusions contained herein are those of the authors
and should not be interpreted as necessarily representing the official
policies or endorsements, either expressed or implied, of any of the
above organizations or any person connected with them.
}}
\author{Kristina Lerman~~~~Aram Galstyan\\USC Information Sciences Institute \and Tad
Hogg\\Hewlett-Packard Labs}

\comment{\author{Kristina \surname{Lerman}$^1$
\email{lerman@isi.edu}}
\author{Aram \surname{Galstyan}$^1$ \email{galstyan@isi.edu}}
\author{Tad \surname{Hogg}$^2$ \email{tad.hogg@hp.com}}
\runningauthor{Kristina Lerman, Aram Galstyan and Tad Hogg}
\runningtitle{Mathematical Analysis of Multi-Agent Systems}
\institute{1. USC Information Sciences Institute\\
2. Hewlett-Packard Laboratories} }
\date{}
\maketitle

\begin{abstract}
We review existing approaches to mathematical modeling and
analysis of multi-agent systems in which complex collective
behavior arises out of local interactions between many simple
agents. Though the behavior of an individual agent can be
considered to be stochastic and unpredictable, the collective
behavior of such systems can have a simple probabilistic
description. We show that a class of mathematical models that
describe the dynamics of collective behavior of multi-agent
systems can be written down from the details of the individual
agent controller. The models are valid for Markov or memoryless
agents, in which each agents future state depends only on its
present state and not any of the past states. We illustrate the
approach by analyzing in detail applications from the robotics
domain: collaboration and foraging in groups of robots.

\vspace{1cm} \noindent {\bf Keywords:} multi-agent systems,
robotics, mathematical analysis, stochastic systems
\end{abstract}


\newpage

\section{Introduction}
\label{sec-Intro}

Distributed systems composed of large numbers of relatively simple
autonomous agents are receiving increasing amount of attention in
the Artificial Intelligence (AI), robotics and networking
communities. Unlike complex deliberative agents, the subject of
much of AI research of the past two decades, simple agents have
no, or limited, capacity to reason about data, plan action or
negotiate with other agents. Although individual agents are far
less powerful than traditional deliberative agents, distributed
multi-agent systems based on such simple agents offer several
advantages over traditional approaches: specifically robustness,
flexibility, and scalability. Simple agents are less likely to
fail than more complex ones. If they do fail, they can be pulled
out entirely or replaced without significantly impacting the
overall performance of the system. They are, therefore, tolerant
of agent error and failure. They are also highly scalable –--
increasing the number of agents or task size does not require
changes in the
agent control programs nor compromise the performance of the
system. In systems using deliberative agents, on the other hand,
the high communications and computational costs required to
coordinate group behavior limit the size of the system to at most
a few dozen agents. Larger versions of such systems require
division into subgroups with limited and simplified interactions
between the groups~\cite{simon96}. In many cases, these
interacting subgroups can in turn be viewed abstractly as agents
following relatively simple protocols, as, for example, in
market-based approaches to multi-agent
systems~\cite{clearwater96}.

There is no central controller directing agents' behavior, rather,
these multi-agent systems are {\em self-organizing}, meaning that
constructive {\em collective} ( macro\-scopic) behavior emerges
from {\em individual} (micro\-scopic) decisions agents make. In
most cases, these decisions are based on purely local information
(that comes from other agents as well as the environment).
Self-organization is ubiquitous in nature --- bacteria colonies,
amoebas and social insects such as ants, bees, wasps, termites,
among others --- display interesting manifestations of this
phenomenon. Indeed, in many of these systems while the individual
and its behavior appear simple to an outside observer, the
collective behavior of the colony can often be quite complex. The
apparent success of these organisms has inspired computer
scientists and engineers to design algorithms and distributed
problem-solving systems modeled after them ({\em e.g.}, swarm
intelligence~\cite{Beni88,BDT99,SchoonderwoerdEtAl97} and
biologically-inspired
systems~\cite{Melhuish98,GolMat00,IMB2001,bojinov02}). Moreover,
current developments in micromechanical systems
(MEMS)~\cite{berlin97} and proposals for coordinated behavior
among microscopic robots~\cite{abelson99,freitas99} will require
agent control programs capable of scaling to extremely large
numbers of agents. Such agents will encounter various microscopic
environments, some with counterintuitive
properties~\cite{purcell77}, making it difficult to design
appropriate deliberative control programs. Furthermore, at least
in their initial development, such machines are likely to face
severe power, computational and communication limitations and
hence require a focus on collective behavior from computationally
simple agents.

\comment{ Two paradigms dominate the design of multi-agent
systems. The first, what we will call the traditional paradigm, is
based on deliberative agents, while the second, the swarm
paradigm, is based on simple agents and distributed control. In
the past two decades, researchers in the Artificial Intelligence
and related communities have, for the most part, operated within
the traditional paradigm. They focused on making the individual
agents, be they software agents or robots, smarter and more
complex by giving them the ability to reason, negotiate and plan
action. In these deliberative systems complex tasks can be done
either individually or collectively. If collective action is
required to complete some task, a central controller is sometimes
used to coordinate group behavior. The controller keeps track of
the capabilities and state of each agent, decides which agents are
best suited for a specific task, assigns them to the agents and
coordinates communication between them (see, {\em e.g.}, Electric
Elves \cite{Elves}, RETSINA \cite{RETSINA}). Deliberative agents
are also capable of collective action in the absence of central
control; however, in these cases agents require knowledge about
capabilities and states of other agents with which they may
interact. For instance, a multi-agent system may break into a
number of coalitions containing several agents, with each
coalition being able to accomplish some task more effectively than
a single agent can. One approach to coalition formation is for
agents to compute the optimal coalition structure
\cite{SandholmShehory:99} and form coalitions based on this
calculation.  Acquiring the knowledge necessary to coordinate
collective behavior may be expensive and impractical, especially
for systems containing many agents.

Swarm Intelligence~\cite{Beni88,BenWan89,BDT99} represents an
alternative approach to the design of multi-agent systems. Swarms
are composed of many simple agents. There is no central controller
directing the behavior of the swarm, rather, these systems are
{\em self-organizing}, meaning that constructive collective ({\em
macro\-scopic}) behavior emerges from local ({\em micro\-scopic})
interactions among agents and between agents and the environment.
Self-organization is ubiquitous in nature --- bacteria colonies,
amoebas and social insects such as ants, bees, wasps, termites,
among others --- are all examples of this phenomenon. Indeed, in
many of these systems while the individual and its behavior appear
simple to an outside observer, the collective behavior of the
colony can often be quite complex. One of the more fascinating
examples of self-organization is provided by {\em Dictyostelium
discoideum}, a species of soil amoebas. Under normal conditions of
abundant food supply these single-celled organisms are solitary
creatures. However, when food becomes scarce, the amoebas emit
chemical signals which cause the colony to aggregate, creating a
multicelled organism \cite{devreotes89dictyostelium,Loomis99} that
is sensitive to light and heat and can travel far more efficiently
than its component cells~\cite{NYTimes}. Next, the collections of
cells within the super-organism differentiate into two distinct
types --- stalks and spores --- which allows the colony to
reproduce by releasing spores which will lie dormant, waiting for
more favorable conditions. There are multiple examples of complex
collective behavior among social insects as well: trail formation
in ants, hive building by bees and mound construction by termites
are just few of the examples. The apparent success of these
organisms has inspired computer scientists and engineers to design
algorithms and distributed problem-solving systems modeled after
them \cite{BDT99,GolMat00,IMB2001,SchoonderwoerdEtAl97}.

Swarms offer several advantages over traditional systems based on
deliberative agents: specifically robustness, flexibility,
scalability, and suitability for analysis. Simple agents are less
likely to fail than more complex ones. If they do fail, they can
be pulled out entirely or replaced without significantly impacting
the overall performance of the system. They are, therefore,
tolerant of agent error and failure. Swarm-based systems are also
highly scalable –-- increasing the number of agents or task size
does not greatly affect system's performance. In systems using
deliberative agents, the high communications and computational
costs required to coordinate agent behavior limit the size of the
system to at most a few dozen agents. Finally, the simplicity of
agent's interactions with other agents make swarms amenable to
quantitative mathematical analysis. }

The main difficulty in designing multi-agent systems (MAS) with
desirable self-organized behavior is understanding the effect
individual agent characteristics have on the collective behavior
of the system. In the past, few analysis tools have been available
to researchers, and it is precisely the lack of such tools that
has been a chief impediment to the wider deployment of
biologically-inspired MAS. Another impediment has been the
difficultly of building hardware required for large-scale
experiments. Researchers had a choice of experiments with
relatively few agents or simulation for studying behavior of a
MAS. Experiments with real agents, {\em e.g.}, robots, allow them
to observe MAS under real conditions; however, experiments are
very costly and time consuming, and systematically varying
individual agent parameters to study their effect on the group
behavior is often impractical. Simulations, such as sensor-based
simulations of robots~\cite{StagePlayer,Michel98},  attempt to
realistically model the environment, the robots' imperfect sensing
of and interactions with it. Though simulations are much faster
and less costly than experiments, they suffer from many of the
same limitations, namely, they are tedious and the results are not
generalizable. Exhaustive scan of the entire parameter space is
often required to reach any conclusion. Moreover, simulations do
not scale well with the system size --- unless computation is
performed in parallel, the greater the number of agents, the
longer it takes to obtain results.

Mathematical modeling and analysis offer an alternative to the
time-consuming and costly experiments and simulations. Using
mathematical analysis we can study dynamics of multi-agent
systems, predict long term behavior of even very large systems,
gain insight into system design: {\em e.g.}, what parameters
determine group behavior and how individual agent characteristics
affect the MAS. Additionally, mathematical analysis may be used to
select parameters that optimize group performance, prevent
instabilities, \etc. Conversely, such analytical tools can also
provide design guidelines for agent programs. Specifically, these
tools rely on various simplifying approximations to the agent
behaviors. By deliberately designing agents to closely match these
approximations, the resulting collective behavior will correspond
to the analytic predictions. In this case, the tools will give a
good indication of how to optimize the design to achieve desired
behaviors. As one example, in market-based
systems~\cite{clearwater96}, designing agents to satisfy the
assumptions of purely competitive markets allows a simple analysis
of resulting behaviors in terms of market equilibria. Of course,
restricting the agent design choices to achieve a close
correspondence with analytic tools may limit the performance of
the system, but this may be a useful tradeoff to achieve a simpler
understanding of overall system behavior. Thus an important
question for applying mathematical analysis for multi-agent
systems is identifying situations in which simple analytic tools
give a useful approximation to the collective behavior.

Mathematical modeling and analysis of large-scale collective
behaviors is being increasingly  used outside of the physical
sciences where it has had much success. It has been applied to
ecology~\cite{ecology}, epidemiology~\cite{epidemiology}, social
dynamics~\cite{Helbing}, artificial
intelligence~\cite{hogg87PhysRep}, and behavior of
markets~\cite{hirshleifer78}, to name just a few disciplines.

In this paper we survey existing work on mathematical modeling and
analysis of artificial multi-agent systems. We also describe a
methodology for creating analytic models of collective behavior of
a multi-agent system. This type of analysis is valid for systems
composed of agents that obey the Markov property: where each
agent's future state depends only on its present state. Many of
the currently implemented multi-agent systems, specifically,
reactive and behavior-based robotics, satisfy this property.
\comment{Our approach is based on viewing these systems as
stochastic systems. We derive a class of mathematical models that
describe the collective dynamics of a multi-agent system. The
resulting models are often quite simple and may be easily written
down by analyzing the behavior of a single agent. Our approach is
general and applicable to many kinds of agent-based systems, such
as software agents~\cite{HubermanHogg88,Lerman00a},
robots~\cite{Lerman01a,Lerman02a} and sensors.} We illustrate the
approaches on robotics problems, comparing theoretical predictions
with experimental and simulations results whenever possible.

\section{Mathematical Models}
\label{sec-Math}

A mathematical model is an idealized representation of a process.
Constructing a mathematical model proceeds incrementally. To be
useful, the model must explicitly include the salient details of
the process it describes so its predictions reasonably match the
actual behaviors of interest. On the other hand, the model should
also be as simple as possible, ideally to allow analytic treatment
and identification of qualitatively important relationships
between individual and system behaviors. The precise choice of
model involves a tradeoff between accuracy in describing reality
and ease of use in providing explanations of the behavior. In our
analysis we will strive to construct the simplest mathematical
model that captures all of the most important details of the
multi-agent system we are trying to describe.

Mathematical models can generally be broken into two classes:
$mic\-ro\-sco\-pic$ and $macro\-sco\-pic$. Microscopic
descriptions treat the agent as the fundamental unit of the model.
There are several variations of the microscopic approach, as
described in the following section. Macroscopic models, on the
other hand, directly describe the collective behavior of a group
of agents. Such models are the focus of the present paper.

\subsection{Microscopic Models}
Microscopic models treat the individual agent as the fundamental
unit of the model. These models describe the agent's interactions
with other agents and the environment. Solving or simulating a
system composed of many such agents gives researchers an
understanding of the global behavior of the system.

\subsubsection{Equations of Motion Approach}
A common method used by physicists to study a system consisting of
multiple entities consists of writing down and solving equations
of motion for each entity. This approach has been adapted by some
to describe agent-based pattern forming systems~\cite{Bonabeau97},
including behaviors exhibited by colonies of biological organisms,
such as slime mold~\cite{Loomis99} and social
insects~\cite{Deneubourg92}. Of particular relevance to the agents
community is the work of Schweitzer and
coworkers~\cite{Schweitzer+al97,HelbingSchweitzer} on the active
walker model of trail formation by ants and people. Active walkers
are randomly walking agents that can influence the environment
(\eg, by depositing pheromone), in addition to being influenced by
it. Schweitzer \ea\ proposed a microscopic model of the
interaction of the ground potential (created by pheromone or
pedestrians' footprints) with the equations of motion of active
walkers.

For large systems, solving equations with many degrees of freedom
is often impractical. In some cases, it may be possible to derive
a macroscopic model with fewer degrees of freedom from the
microscopic model. Helbing, Schweitzer and coworkers did this in a
later work~\cite{HelbingSchweitzer}, where they derived a model
that describes the behavior of subpopulations of active walkers.
Although these models of trail formation may be faulted for not
being biologically realistic, they reproduce trail-forming
behavior of real ants, such as the ability to discover and link
distributed food sources without {\em a priori} knowledge of their
location. Such models may be especially useful to study
pheromone-based trail formation and navigation in
robots~\cite{VauStoSukMat00,VauStoSukMat00b}.

The main disadvantages of microscopic models are their poor
scaling properties and that it is not always easy or obvious how
to write down the equations of motion of each agent. Even if a
model can be written down, in most cases it will not be
analytically tractable, and the solution will have to be simulated
on a computer. By resorting to simulation, one looses much of the
power of mathematical analysis.

\subsubsection{Microscopic Simulations}
Microscopic simulations, such as molecular
dynamics~\cite{RobDuncan}, cellular
automata~\cite{abelson99,Wolfram} and particle hopping
models~\cite{CSS00}, are a popular tool for studying dynamics of
large multi-agent systems. In these simulations, agents change
state stochastically or depending on the state of their neighbors.
The popular Game of Life is an example of cellular automata
simulation. Another example of the microscopic approach is the
probabilistic model developed by Martinoli and
coworkers~\cite{Martinoli99,MarIjsGam99,IMB2001} to study
collective behavior of a group of robots. Rather than compute the
exact trajectories and sensory information of individual robots,
Martinoli {\em et al.} model each robot's interactions with other
robots and the environment as a series of stochastic events, with
probabilities determined by simple geometric considerations.
Running several series of stochastic events in parallel, one for
each robot, allowed them to study the group behavior of the
multi-robot system.

\subsection{Macroscopic Models}
A macroscopic description offers several advantages over the
microscopic approach. It is more computationally efficient,
because it uses many fewer variables. A macroscopic model can
often be solved analytically, yielding important insights into the
behavior of quantities of interest. The macroscopic descriptions
also tend to be more universal, meaning the same mathematical
description can be applied to other systems governed by the same
abstract principles. At the heart of this argument is the concept
of separation of scales, which holds that the details of
microscopic interactions (among agents) are only relevant for
computing the values of the parameters of the macroscopic model.
This idea has been used by physicists to construct a single model
that describes the behavior of seemingly disparate systems, {\em
e.g.}, pattern formation in convecting fluids and chemical
reaction-diffusion systems~\cite{Walgraef97}. This principle of
systems consisting of nearly decomposable parts applies broadly
not only to physical systems but also to naturally evolved
systems, as found in biology and economics, and designed
technological artifacts~\cite{courtois85,simon61,simon96}. From
the perspective of large-scale agent systems, this decomposition
often arises from processing, sensory and communication
limitations of the individual agents. In effect, these limits mean
agents can only pay attention to a relatively small number of
variables in the full system~\cite{hogg87PhysRep}, and will
generally communicate concise summaries of their states to other
agents. Of course, the two description levels are related, and it
may be possible in some cases to derive the parameters of the
macroscopic model from microscopic theory.

Macroscopic models are very popular and have been successfully
applied to a wide variety of problems in physics, chemistry,
biology and the social sciences. In most of these applications,
the microscopic behavior of individual entity (a Brownian particle
in a volume of gas or an individual residing in US) is quite
complex, often stochastic and unpredictable, and certainly
analytically intractable. Rather than account for the inherent
variability of individuals, scientists model the behavior of some
{\em average} quantity that represents the system they are
studying (volume of gas or population of US). Such macroscopic
descriptions often have a very simple form and are analytically
tractable. They are sometimes called phenomenological models,
because they are not derived from microscopic theories. It is
important to remember that such models do not reproduce the
results of a single experiment --- rather, the behavior of some
observable averaged over many experiments or observations. Such a
probabilistic approach is the basis for statistical physics.

Usually, the relative size of fluctuations in statistical systems
decreases with the number of components. In these cases, the
system is almost always found near its average behavior and so the
average is a good description for most individual experiments. It
is this observation that allows the convenient study of average
properties to describe behaviors actually seen in most
experiments. In some cases, fluctuations can become large, \eg,
near phase transitions in physical systems. Such behaviors are
also seen in computational systems, particularly those involved
with combinatorial search~\cite{hogg96d}, where long-tailed
distributions have typical behavior far from that of the average.
Thus when using macroscopic models to determine behavior of
averages, it is important to keep in mind the possibility that
actual system behaviors could be far from average. Fortunately, in
the context of multi-agent systems, such large fluctuations will
require an unexpectedly large statistical correlation in agent
behaviors, which is unlikely in situations in which the agents are
fairly independent and each act on only a few aspects of the
overall system state (\eg, based on local sensory information).

\subsubsection{Finite Difference Equations}
A macroscopic model can be frequently written down as a finite
difference equation describing the change in the value of a
dynamic variable over some time interval $\Delta t$. For example,
in a model of population dynamics of US,
\begin{eqnarray*}
  N(t+\Delta t)& =& N(t)+\Delta t R(t)N(t)\\
  R(t) & = & \frac {N(t+\Delta t) - N(t)} {\Delta t N(t)},
\end{eqnarray*}
\noindent where $N(t)$ is the (time-dependent) US population,
$\Delta t$ is a decade used by the Census Bureau, and $R(t)$ it
the rate of change of population due to births, deaths,
immigration and emigration. In general, $R(t)$ will also depend on
the choice of $\Delta t$. The modeler finds an appropriate $R$ to
describe population growth of US, and solves the equations to
project population growth into the future.

This description provides a finite-difference equation for the
behavior of $N$ at integer multiples of $\Delta t$. This has been
used to model a robotic
system~\cite{MartinoliISER02,MartinoliEaston02}, and is
particularly appropriate for \emph{synchronous} systems, \ie,
where all agents make decisions at the same time (such as parallel
update cellular automata).

In the continuous limit ($\Delta t \rightarrow 0$ or large $N$),
the difference equation become a differential equation, known as
the rate equation. For the example above, the rate equation is
 $ \frac {dN(t)}{dt}=R(t)N(t)$.
which is also applicable to \emph{asynchronous} systems. In many
cases, the behavior of this differential equation matches that of
the difference equation~\cite{bender78}. However, this is not
always the case: synchronous and asynchonous systems can have very
different collective behaviors~\cite{huberman93a}. Large scale
agent systems interacting with an environment, \eg, robots, will
often need to respond to environmental signals that arrive at
unpredictable times. Such systems are likely to be better viewed
as asynchronous, for which the differential equation approach is
most suited.

\subsubsection{Rate Equations}
An alternate way to derive the rate equation is to start with the
master equation for a stochastic system and macroscopically
average it to get the rate equation for the dynamics of average
quantities. Section~\ref{sec:Math2} presents a tutorial on this
approach. However, in order to create a model of a multi-agent
system, one does not need to start with the master equation ---
one can easily write down the rate equations by examining the
details of individual agent controller.

The rate equations are deterministic. In stochastic systems,
however, rate equations describe the dynamics of average
quantities. How closely the average quantities track the behavior
of the actual dynamic variables depends on the magnitude of
fluctuations. Usually the larger the system, the smaller are the
(relative) fluctuations. In a small system, the experiment may be
repeated many times to average out the effect of fluctuations.
Pacala {\em et al.}\cite{Pacala} showed that in models of task
allocation in ants, the exact stochastic and the average
deterministic models {\em quantitatively} agree in systems
containing as few as ten ants. The agreement increases as the size
of the system grows. Martinoli and Easton~\cite{MartinoliISER02}
have shown quantitative agreement with simulations in a system of
16--24 robots.

The rate equation has been used to model dynamic processes in a
wide variety of systems. The following is a short list of
applications: in chemistry, it has been used to study chemical
reactions~\cite{Gardiner}; in physics, the growth of semiconductor
surfaces \cite{BarabasiStanley95} among others; in ecology to
study dynamics of populations~\cite{Cushing}, including
predator-prey systems \cite{Haberman}; in biology to model the
behavior of social insects \cite{Pacala,Theraulaz98}. The rate
equation has also found application in the social sciences
\cite{Helbing} and in AI. Huberman, Hogg and
coworkers~\cite{HubermanHogg88,Kephart90} mathematically studied
collective behavior of a system of agents (they called a
computational ecology), where each agent chooses between two
alternative strategies. They start with underlying probability
distributions and derive rate equations for the average numbers of
agents using each strategy. In fact, the same equations can be
written down by examining the macroscopic state diagram of the
agents. Yet another application of the approach presented here is
coalition formation in electronic marketplaces~\cite{Lerman00a}.

In the robotics domain, Sugawara and
coworkers~\cite{Sugawara97,SugSanYosAbe98} developed simple
analytical models of cooperative foraging in groups of
communicating  and non-co\-mmu\-ni\-cating robots. \comment{We
discuss their models in Section~\ref{sec:foraging-sugawara}.
Although these models are similar to ours, they are overly
simplified and fail to take crucial interactions among robots into
account.} Kazadi \ea~\cite{Kazadi02} study the general properties
of multi-robot aggregation using phenomenological macroscopic
models. Agassounon and Martinoli~\cite{Agassounon02} present a
model of aggregation in which the number of robots taking part in
the clustering task is based on the division of labor mechanism in
ants. Lerman \emph{et al.} have analyzed
collaborative~\cite{Lerman01a} and foraging~\cite{Lerman02a}
behavior in groups of robots. Results from these works will be
used to illustrate the modeling methodology described in this
paper. The focus of the present paper is to show that there is a
principled way to construct a macroscopic analytic model of
collective dynamics of a MAS, and, more importantly, a practical
``recipe'' for creating such a model from the details of the
microscopic agent controller.

\section{Macroscopic Analytic Models}
\label{sec:Math}
\subsection{MAS as Stochastic Systems}
The behavior of individual agents in a multi-agent system has many
complex influences, even in a controlled laboratory setting.
Agents are influenced by external forces, many of which may not be
anticipated. For robots, external forces include friction, which
may vary with the type of surface the robot is moving on, battery
power, sound or light signals, {\em etc}. Even if all the forces
are known in advance, the agents are still subject to random
events: fluctuations in the environment, as well as noise in the
robot's sensors and actuators. Each agent will interact with other
agents that are influenced by these and other events. In most
cases it is difficult to predict the agents' exact trajectories
and thus know which agents will come in contact with one another.
Finally, the agent designer can take advantage of the
unpredictability and incorporate it directly into the agent's
behavior. For example, the simplest effective policy for obstacle
avoidance in a robot is for it to turn a random angle and move
forward. In summary, the behavior of simple agents in a
complicated environment is so complex, the MAS is best described
probabilistically, as a stochastic system.

Before we present a methodology for mathematical a\-na\-ly\-sis of
sto\-chas\-tic sys\-tems, we need to define some terms. {\em
State} labels a set of related agent behaviors required to
accomplish a task. For example, when a robot is engaged in a
foraging task, its goal is to collect objects, such as pucks,
scattered around the arena and bring them to a home base. The
foraging task can be thought of as consisting of the following
high-level behavioral requirements~\cite{Arkin} or states
\begin{description}
\item[Homing] --- return the puck to a home base after it is picked up (includes collision avoidance)
\item[Pickup] --- if a puck is detected, close gripper
\item[Searching] --- wander around the arena in search of pucks (includes collision avoidance)
\end{description}
Each of these high level states may consist of a single action or
behavior, or a set of behaviors. For example, when a robot is in
the {\em Sear\-ching} state, it is wandering around the arena,
detecting objects and avoiding obstacles. In the course of
accomplishing the task, the robot will transition from the {\em
Searching} to {\em Pickup} and finally to {\em Homing} states. We
define these states to be mutually exclusive, \ie, so that each
agent in a multi-agent system is in exactly one of a finite number
of states during a sufficiently short time interval. Note that
there can be one-to-one correspondence between agent
actions/behaviors and states. However, in order to keep the
mathematical model compact and tractable, it is useful to {\em
coarse-grain} the system by choosing a smaller number of states,
each incorporating a set of agent actions or behaviors. Such
coarse-graining is particularly relevant when we are only
interested in behaviors described at this coarser level of
abstraction.

In general, the full description of an agent in its environment
could involve an arbitrary amount of detail, \eg, its exact
location in the environment. While such continuous states could be
included in the formalism we present here, for simplicity we take
the possible states of interest to be a finite set. Even when
continuous values, such as location, are relevant to particular
applications, it may be sufficient to treat these values with just
a few coarse-grained regions.

We associate a unit vector $\hat{q}_k$ with each state $k =
1,2,\ldots,L$. The configuration of the system is defined by the
occupation vector
\begin{eqnarray}
\label{eqn-occupation} {\vec n} = \sum_{k=1}^{L}{n_k {\hat q_k}}
\end{eqnarray}
where $n_k$ is the number of agents in state $k$. The probability
distribution $P({\vec n},t)$ is the probability the system is in
configuration ${\vec n}$ at time $t$.

\subsection{The Stochastic Master Equation: A Tutorial}
\label{sec:Math2} For systems that obey the Markov property, the
future is determined only by the present and not by the past.
Clearly, agents that plan or use memory of past actions to make
decisions, will not meet this criterion directly. While it is
always possible to include the contents of an agent's memory as
part of its state to maintain the Markov property, this can result
in a vast expansion in the number of states to consider.

Fortunately, many MAS studied by various researchers, specifically
those based on reactive and behavior-based robots and many types
of software agents and sensors, do satisfy the Markov property
with a fairly modest number of states. We restate the Markov
property in the following way: the configuration of a system at
time $t + \Delta t$ depends only on the configuration of the
system at time $t$. In terms of coarse-grained states, we require
this property to apply to the system described at this level of
abstraction, at least to sufficient approximation.

The Markov property allows us to rewrite the marginal probability
density $P({\vec n}, t + \Delta t)$ in terms of conditional
probabilities for transition from ${\vec{n}}'$ to ${\vec n}$:
\begin{eqnarray*}
P({\vec n}, t + \Delta t) = \sum_{{\vec{n}}'}{P({\vec n}, t +
\Delta t |{{\vec{n}}'},t)P({{\vec {n}}'}, t)}.
\end{eqnarray*}
Using the fact that
\begin{eqnarray*}
\sum_{{\vec{n}}'}{P({{\vec {n}}'}, t + \Delta t|{\vec n},t)} = 1,
\end{eqnarray*}
allows us to write the change in probability density as
\begin{eqnarray}
\label{eqn-probability} P({\vec n}, t + \Delta t) - P({\vec n}, t)
&=& \sum_{{{\vec{n}}}'}{P({\vec n}, t + \Delta t |{\vec
{n}}',t)P({\vec {n}}', t)} \nonumber \\
& & - \sum_{\vec{n}'}{P({\vec {n}}', t + \Delta t |{\vec
n},t)P({\vec n}, t)}.
\end{eqnarray}

\noindent In the continuum limit, as ${\Delta t} \rightarrow 0$,
Eq.~\ref{eqn-probability} becomes
\begin{eqnarray}
\label{eqn-master} {{\partial P({\vec n}, t)}\over {\partial t}} =
\sum_{\vec{n}'}{W({\vec n}|{\vec{n}'};t)P({\vec{n}}', t)} -
\sum_{\vec{n}'}{W({\vec {n}'}|{\vec n};t)P({\vec n}, t)}\,,
\end{eqnarray}
with transition rates defined as
\begin{eqnarray}
\label{eqn-transition} W({\vec n}|{\vec {n}'};t) = {\lim_{\Delta t
\rightarrow 0}} {{P({\vec n}, t + \Delta t |{\vec {n}'},t)} \over
{\Delta t}},
\end{eqnarray}
provided this limit exists, \eg, changes are not synchronized to a
global clock which would make $P({\vec n}, t + \Delta t |{\vec
{n}'},t)$ equal to zero for all ${\vec n} \neq {\vec {n}'}$ when
$\Delta t$ is sufficiently small.

Equation \ref{eqn-master} says that the configuration of the
system is changed by transitions to and from states. It is known
as the Master Equation and is used widely to study dynamics of
stochastic systems in physics and chemistry~\cite{Gardiner},
traffic flow~\cite{Mahnke97,Mahnke99} and
sociodynamics~\cite{Helbing}, among others. The Master Equation
also applies to semi-Markov processes in which the future
configuration depends not only on the present configuration, but
also on the time the system has spent in this configuration. As we
will see in a later section,  transition rates in these systems
are time dependent, while in pure Markov systems, they are
time-independent.

\subsection{The Rate Equation}

The Master equation (Eq.~\ref{eqn-master}) fully determines the
evolution of a stochastic system. Once the probability
distribution $P({\vec {n}}, t)$ is found, one can calculate the
characteristics of the system, such as the average and the
variance of the occupation numbers. The Master equation is almost
always too complex to be analytically tractable. Fortunately we
can vastly simplify the problem by working with the average
occupation number, $\langle \vec{n} \rangle$ (the Rate Equation).
To derive an equation for $\langle \vec{n} \rangle$, we multiply
Eq.~\ref{eqn-master} by $ \vec{n}$ and take the sum over all
configurations:
\begin{eqnarray}
\label{eqn-rate-1}
 {\partial \over {\partial t}}\langle \vec{n} \rangle
& \equiv & {\partial \over {\partial t}}\sum_{\vec{n}}{\vec{n}
P({\vec{n}},t)} \nonumber \\
& = & \sum_{\vec{n}} \sum_{\vec{n}'}\vec{n}{W({\vec n}|{\vec
{n}'};t)}P(\vec{n}',t)
 -\sum_{\vec{n}}\sum_{\vec{n}'}\vec{n}{W({\vec n}'|{\vec
{n}};t)}P(\vec{n},t) \nonumber \\
& = & \sum_{\vec{n}} \sum_{\vec{n}'}(\vec{n}'-\vec{n}){W({\vec
n}'|{\vec{n}};t)}P(\vec{n},t) \nonumber \\
& =& \biggl < \sum_{\vec{n}'}(\vec{n}'-\vec{n}){W({\vec
n}'|{\vec{n}};t)} \biggl
>
\end{eqnarray}
where $\langle ... \rangle$ stands for averaging over the
distribution function $P(\vec n,t)$.  The time--evolution of a
particular occupation number is obtained from the vector equation
Eq.~\ref{eqn-rate-1} as

\begin{equation}
\label{eqn-rate-2} {\partial \over {\partial t}}\langle n_k
\rangle=\biggl < \sum_{\vec{n}'}(n_{k}'-n_k){W({\vec
n}'|{\vec{n}};t)} \biggr >
\end{equation}

Let us assume for simplicity that only individual transitions
between states are allowed,
 {\em i.e.}, $W({\vec n}'|{\vec{n}};t)\neq 0$ only if ${\vec n}'-{\vec
 n}=\hat{q}_i-\hat{q}_j$, $i\neq j$,
 and let $w_{ij}$ be the transition rate  from
state $j$ to state $i$. This is appropriate for systems with
asynchronous updates for the agents since in that case it is very
unlikely that two agents would make a change at the same time.
Note that in general $w_{ij}$ may be a function of the occupation
vector $\vec{n}$, $w_{ij}=w_{ij}(\vec{n})$. Define a matrix ${\bf
D}$ with off-diagonal elements $w_{ij}$ and with diagonal elements
$D_{ii}=-\sum_kw_{ki}$. Then we can rewrite Eq.~\ref{eqn-rate-1}
in a matrix form as
\begin{equation}
{\partial \over {\partial t}}\langle \vec{n} \rangle=\langle {\bf
D}(\vec n ) \cdot \vec n \rangle \approx {\bf D}(\langle \vec n
\rangle)\cdot \langle \vec n \rangle
\end{equation}
where we have used so the called mean-field approximation $\langle
F(\vec{n})\rangle \approx F(\langle \vec n \rangle)$. The
mean-field approximation is often used in statistical phy\-sics
and is well justified for unimodal and sharp distribution
functions~\cite{opper01}.
The average occupation numbers obey the following system of
coupled linear equations

\begin{equation}
\label{eqn-rate-3} {\partial \over {\partial t}}\langle n_k
\rangle= \sum_{j} w_{jk}(\langle \vec n \rangle) \langle
n_{j}\rangle - \langle n_k \rangle \sum_{j}w_{kj}(\langle \vec n
\rangle)
\end{equation}

The above equation is known as the Rate Equation. It has the
following interpretation: occupation number $n_k$ will increase in
time (first term in Eq.~\ref{eqn-rate-3}) due to transitions from
other states to state $k$, and it will decrease in time due to the
transitions from the state $k$ to other states (second term).

\subsubsection {Transition rates}

Finding an appropriate mathematical form for the transition rates
is the main challenge in applying the rate equations to real
systems. Usually, the transition is triggered when an agent
encounters some stimulus
--- be it another agent in a particular state, an object, its
location, {\em etc}. For simplicity, we will assume that agents
and triggers are uniformly distributed in space (though we will
consider systems where agents interact in space, it does not
necessarily have to be physical space, but a network, the Web,
{\em etc}). The assumption of spatial uniformity may be reasonable
for agents that randomly explore space ({\em e.g.}, searching
behavior in robots tends to smooth out any inhomogeneities in the
robots' initial distribution); however, it fails for systems that
are strongly localized, for instance, where all the objects to be
collected by robots are located in the center of the arena. In
these anomalous cases, the transition rates will have a more
complicated form and in some cases it may not be possible to
express them analytically altogether. If the transition rates
cannot be calculated from first principles, it may be expedient to
leave them as parameters in the model and estimate them by fitting
the model to data.


\subsubsection{Rate Equation and Multi-Agent Systems}
The rate equation is a useful tool for mathematical analysis of
macroscopic, or collective, dynamics of many agent-based systems.
To facilitate the analysis, we begin by drawing the macroscopic
state diagram of the system. The state diagram can be constructed
from the details of the individual agent's behavior, as will be
illustrated in the applications. Not every microscopic, or
individual agent, behavior need become a macroscopic state. In
order to keep the model tractable, it is often useful to
coarse-grain the system by considering several related behaviors
as a single state. As as example considered in detail below, we
may take the searching state of robots to consist of the actions
{\em wander in the arena, detect objects} and {\em avoid
obstacles}. When necessary, the searching state can be split into
three states, one for each behavior; however, we are often
interested in the $minimal$ model that captures the important
behavior of the system. Coarse-graining presents a way to
construct such a minimal model. In addition to states, we must
also specify transitions between states. These will be represented
as arrows leading from one state to another.

Each state in the macroscopic state diagram corresponds to a
dynamic variable in the mathematical model --- the average number
of agents in that state --- and it is coupled to other variables
via transitions between states. The mathematical model will
consist of a series of coupled rate equations, one for each state,
which describes how the number of agents in that state changes in
time. This quantity may increase due to $incoming$ transitions
from other states, and it may decrease due to $outgoing$
transitions to other states. Every transition will be accounted
for by a term in each equation, with transition rates specified by
the details of the interactions between agents.

In the next sections we will illustrate the details of the
approach by applying it to study problems in the robotics domain.
Our examples include foraging (Section~\ref{sec:foraging}) and
collaboration (Section~\ref{sec:sp}) in groups of robots.

\newcommand{\ti}{{\tilde t}}
\section{Foraging in a Group of Robots}
\label{sec:foraging}

\comment{ Robot collection and foraging is one of the oldest and
most studied problems in robotics. In this task a single robot or
a group of robots has to collect objects scattered around the
arena and to assemble them either in some random location
(collection task) or a pre-specified ``home'' location (foraging
task). These tasks have been studied under a wide variety of
conditions and architectures, both experimentally and in
simulation. Below is a partial list of the previous work on
collection that can be categorized in the following way:
\begin{itm}
\item Task type
    \begin{itm}
    \item collection\cite{Beckers94,MarIjsGam99}
    \item foraging\cite{Mataric92,GolMat00,ArkBalNit}
    \end{itm}
\item System type
    \begin{itm}
    \item single robot
    \item group of robots (homogeneous\cite{GolMat00} and heterogeneous\cite{GolMat00,ParkerPhD})
    \end{itm}
\item Controller type
    \begin{itm}
    \item reactive
    \item behavior-based\cite{Mataric92,GolMat00}
    \item hybrid\cite{ArkBalNit}
    \end{itm}
\item Communication
    \begin{itm}
    \item no communication\cite{GolMat00}
    \item direct\cite{ArkBalNit,Sugawara97}
    \item stigmergetic (through modification of the environment) \cite{HolMel00,VauStoSukMat00}
    \end{itm}
\end{itm}

 The broad appeal of
this problem is explained both by ubiquity of collection in
general and foraging in particular in nature --- as seen in the
food gathering behavior of many insects --- as well as its
relevance to many military and industrial applications, such as
de-mining, mapping and toxic waste clean-up. Foraging has been a
testbed for the design of physical robots and their controllers,
as well as a framework for exploring many issues in the design and
implementation of multi-robot teams.}

Robot collection and foraging are two of the oldest and most
studied problems in robotics. In these tasks a single robot or a
group of robots has to collect objects scattered around the arena
and to assemble them either in some random location (collection
task~\cite{Beckers94,MarIjsGam99}) or a pre-specified ``home''
location (foraging task~\cite{Mataric92,GolMat00,ArkBalNit}).
These tasks have been studied under a wide variety of conditions
and architectures, both experimentally and in simulation: in
homogeneous  and heterogeneous~\cite{GolMat00} systems, using
behavior-based~\cite{Mataric92,GolMat00} and hybrid
control~\cite{ArkBalNit}, no communication~\cite{GolMat00}, direct
communication~\cite{ArkBalNit,Sugawara97}, as well as indirect
communication through the environment~\cite{HolMel00}. The broad
appeal of this problem is explained both by ubiquity of collection
in general and foraging in particular in nature --- as seen in the
food gathering behavior of many insects --- as well as its
relevance to many military and industrial applications, such as
de-mining, mapping and toxic waste clean-up. Foraging has been a
testbed for the design of physical robots and their controllers,
as well as a framework for exploring many issues in the design and
implementation of multi-robot teams.

\begin{figure}
  \epsfxsize = 2.5in
  \center{\epsffile{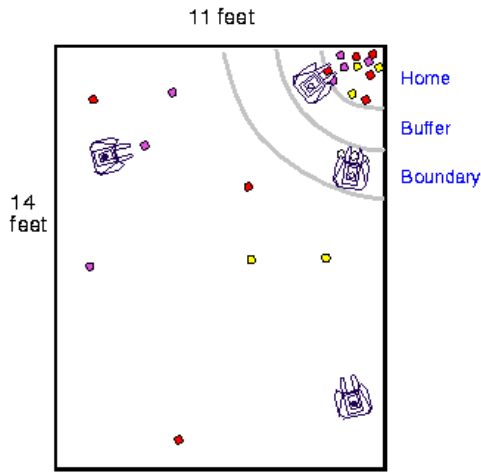}}
  \caption{Diagram of the foraging arena (courtesy of D.~Goldberg).}\label{fig:foraging_robots}
\end{figure}

In this section, we focus on analysis of foraging in a homogeneous
non-communi\-ca\-ting multi-robot systems using behavior-based
control, the type of systems studied by Mataric and collaborators
\cite{Mataric92,GolMat00}. In Sec.~\ref{sec:sugawara} we will
present Sugawara {\em et al.}'s model of foraging in communicating
robots. Figure~\ref{fig:foraging_robots} is a snapshot of a
typical experiment with four robots. The robots' task is to
collect small pucks randomly scattered around the arena. The arena
itself is divided into a search region and a small ``home'', or
goal, region where the collected pucks are deposited. The
``boundary'' and ``buffer'' regions are part of the home region
and are made necessary by limitations in the robots' sensing
capabilities, as described below. Each robot has an identical set
of behaviors governed by the same controller. The behaviors that
arise in the collection task are \cite{GolMat00}:
\begin{description}
  \item[Avoiding] obstacles, including other robots and boundaries.
This behavior is critical to the safety of the robot.
  \item[Searching] for pucks: robot moves forward and at random intervals
  turns left or right through a random arc.  If the robot enters the Boundary region, it
returns to the search region. This prevents the robot from
collecting pucks that have already been delivered.
  \item[Detecting] a puck.
  \item[Grabbing] a puck.
  \item[Homing] if carrying a puck, move towards the home location.
  \item[Creeping] activated by entering Buffer region. The robot will start using the
  close-range detectors at this point to avoid the boundaries.
  \item[Home] robot drops the puck. This activates the exiting behavior.
  \item[Exiting] robot exits the home region and resumes search.
\end{description}

Interference, caused by competition for space between spatially
extended robots, has long been recognized as a critical issue in
multi-robot systems \cite{FonMat96,Sugawara97}. When two robots
find themselves within sensing distance of one another, they will
execute obstacle avoiding maneuvers in order to reduce the risk of
a potentially damaging collision. The robot stops, makes a random
angle turn and moves forward. This behavior takes time to execute;
therefore, avoidance increases the time it takes the robot to find
pucks and deliver them home. Clearly, a single robot working alone
will not experience interference from other robots. However, if a
single robot fails, as is likely in a dynamic, hostile
environment, the collection task will not be completed. A group of
robots, on the other hand, is robust to an individual's failure.
Indeed, many robots may fail but the performance of the group may
be only moderately affected. Many robots working in parallel may
also speed up the collection task. Of course, the larger the
group, the greater the degree of interference --- in the extreme
case of a crowded arena, robots will spend all their time avoiding
other robots and will not bring any pucks home.

Several approaches to minimize interference have been explored,
including communication \cite{Parker98} and cooperative strategies
such as trail formation \cite{VauStoSukMat00b} and bucket brigade
\cite{FonMat96,OstSukMat01}. In some cases, the effectiveness of
the strategy to minimize interference will also depend on the
group size \cite{OstSukMat01}. Therefore, it is important to
quantitatively understand interference between robots and how it
relates to the group and task sizes before choosing alternatives
to the default strategy. For some tasks and a given controller,
there may exist an optimal group size that maximizes the
performance of the system \cite{ArkBalNit,FonMat96,OstSukMat01}.
Beyond this size the adverse effects of interference become more
important than the benefits of increased robustness and
parallelism, and it may become beneficial to choose an alternate
foraging strategy. We will study interference mathematically and
attempt to answer these questions.

Nitz {\em et al.}\cite{ArkBalNit} briefly addressed the question
of what is an appropriate number of robots for a foraging task in
a given environment. By simulating foraging in groups of up to
five communicating robots, they observed an increase in
performance when adding one to three robots as compared to a
single worker. However, the performance seemed to level out and
even degrade with further additions. Performance of
non-communicating robots seemed to improve as the group size grew,
at least up to the group size of five. No simulations for larger
group sizes were carried out.
\comment{Sugawara and coworkers~\cite{Sugawara97,SugSanYosAbe98}
carried out a quantitative study of foraging in groups of
communicating and non-communicating robots. They  developed a
simple rate equation-based model of foraging and analyzed it under
different conditions, that included non-communicating robots.
However, their treatment of interference is different from ours,
as will be explained below. }

\subsection{Rate Equation Model of Foraging}
As mentioned above, interference is the result of competition
between two or more robots for the same resource, be it physical
space, the puck both are trying to pick up, energy, communications
channel, \etc. In the foraging task, competition for physical
space, and the resulting avoidance of collisions with other
robots, is the most common source of interference.  In this
section we examine the foraging scenario where robots are required
to collect pucks and bring them to a specified ``home'' location.

\begin{figure}[htpb]
  \epsfxsize=1.8in
  \center{
  \epsffile{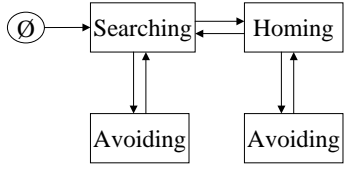}}
  \caption{State diagram of a multi-robot foraging system with homing.}\label{fig:homing_diagram}
\end{figure}

At a macroscopic level, during some short time interval, every
robot can be considered to be in the searching, homing or avoiding
states, as shown in Fig.~\ref{fig:homing_diagram}. We assume that
actions like detecting and grabbing a puck take place on a
sufficiently short time scale that they can be incorporated into
the search state. Likewise, creeping, \etc, can be incorporated
into the homing state.\footnote{If we find that the given
descriptive level does not adequately capture the behavior of a
real or simulated system, we can consider more states in the
model. For now, we are interested in the minimal model that
reproduces salient features of the foraging system.} Initially the
robots are in the search state. When a robot finds a puck, it
picks it up and moves to the ``home'' region. Execution of the
homing behavior requires a period of time $\tau_h$. At the end of
this period, the robot deposits the puck at home and resumes
search for more pucks. While the robot is homing, it will
encounter and try to avoid other robots.

Each state in the diagram corresponds to a dynamic variable. Let
$N_s(t)$, $N_h(t)$, $N_{av}^s(t)$, $N_{av}^h(t)$ be the number of
searching, homing, avoiding while searching and avoiding while
homing robots at time $t$, with the total number of robots,
$N_0=N_s(t)+ N_h(t)+ N_{av}^s(t)+ N_{av}^h(t)$, a constant. We
model the environment by letting $M(t)$ be the number of
undelivered pucks at time $t$.
Also, let $\alpha_r$ be the rate of detecting another robot and
$\alpha_p$ the rate of detecting a puck. These parameters connect
the model to the experiment, and they are related to the size of
the robot and the puck, robot's detection radius and the speed of
the robot. It was shown experimentally~\cite{GolMat00} that
interference is most pronounced near the home region, because the
density of robots is, on average, greater there. Therefore, we
expect the rate of encountering other robots to be greater near
the home region and introduce $\alpha_r^{\prime}$, the rate of
detecting another robot while homing. The following equations
describe the time evolution of the dynamic variables\footnote{For
simplicity, we do not include wall avoidance in the equations, but
do take it into account when fitting model to the data.}:
\begin{eqnarray}
\label{eq:forage-searching3} \frac{d{N_s(t)}}{dt}&=&
-\alpha_pN_s(t)[ M(t) -N_h(t)-N_{av}^h(t) ] \nonumber \\
& & - \alpha_r N_s(t)[N_s(t)+N_0] \nonumber\\
& & + \frac{1}{\tau_h} N_h(t)+\frac{1}{\tau} N_{av}^s(t), \\
\label{eq:forage-homing3} \frac{d{N_h(t)}}{dt}&=&\alpha_pN_s(t)[
M(t) -N_h(t)-N_{av}^h(t) ]  \nonumber
\\
& & -\alpha_r^{\prime}N_h(t)[N_h(t)+N_0]\nonumber \\
& & + \frac{1}{\tau} N_{av}^h(t) -\frac{1}{\tau_h} N_h(t),
\\
\label{eq:forage-avoiding-while-homing3}
\frac{d{N_{av}^h(t)}}{dt}&=&\alpha_r^{\prime}N_h(t)[N_h(t)+N_0]-\frac{1}{\tau}
N_{av}^h(t), \\
\label{eq:forage-pucks3} \frac{d{M}(t)}{d t} & = &
-\frac{1}{\tau_h}N_h(t).
\end{eqnarray}

The first term in Eq.~\ref{eq:forage-searching3} accounts for a
decrease in the number of searching robots when robots find pucks
and start homing. The second term says that the number of
searching robots decreases when two searching robots detect each
other and commence avoiding maneuvers or when a searching robot
detects another robot in any of the remaining states. The number
of available pucks is just the number of pucks in the arena less
the pucks held by the homing robots. The last two terms in the
equation require more explanation. We assume that  it takes on
average $\tau_h$ time for a robot to reach home after grabbing a
puck. Then the average number of robots that deliver pucks during
a short time interval $dt$ and return to the searching state can
be approximated as $dt N_h/\tau_h$. Likewise, after a period of
time $\tau$, $dt N_{av}^s/\tau$ robots leave the avoiding state
and resume searching. We assume that interference will increase
the homing time for each robot: if $\tau^{0}_h$ is the average
homing time in the absence of collisions with other robots, then
the effective homing time can be estimated as
 $ \tau_h = {\tau_h}^0[1 + \alpha_r^{\prime} \tau N_0]\,
$.

The remaining equations have similar interpretations. We can take
advantage of the conservation of the total number of robots to
compute $N_{av}^h(t)$. These equations are solved numerically
under the conditions that initially, at $t=0$, there are $M_0$
pucks and $N_0$ searching robots.

\begin{figure}[htbp]
\center{
  \epsfxsize = 2.4in
  \epsffile{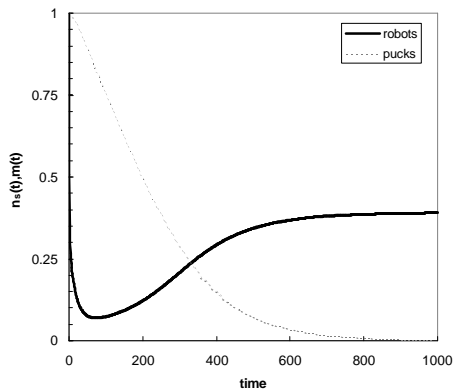}
  \caption{ Time evolution of the fraction of searching robots (solid line) and undelivered pucks (dashed line) for $\tau = 3\ s$, $\alpha_p=0.015$, $\alpha_r=0.04$, and $\alpha_r^{\prime}=0.08$.
} \label{fig:fhom1}
  }
\end{figure}

Figure \ref{fig:fhom1} shows the time evolution of the fraction of
searching robots and pucks for $M_0=20$, $N_0=5$, $\tau= 3\ s$,
$\tau_{h}^{0}=16\ s$. The number of searching robots (solid line)
first quickly decreases as robots find pucks and carry them home,
but then it increases and saturates at some steady state value as
the number of undelivered pucks approaches zero (dashed line). The
fraction of the searching robots in the steady state depends on
the avoiding time parameter, which determines the fraction of
robots in the avoiding state.

To validate results of the model, we ran foraging simulations
using Player/Stage simulator~\cite{StagePlayer} for groups of one
to ten robots and twenty pucks randomly scattered around the
arena. Details of the simulations are presented in
\cite{Lerman02a}.

\begin{figure}[htbp]
\center{
  \begin{tabular}{l}
  \epsfxsize = 2.4in
  \epsffile{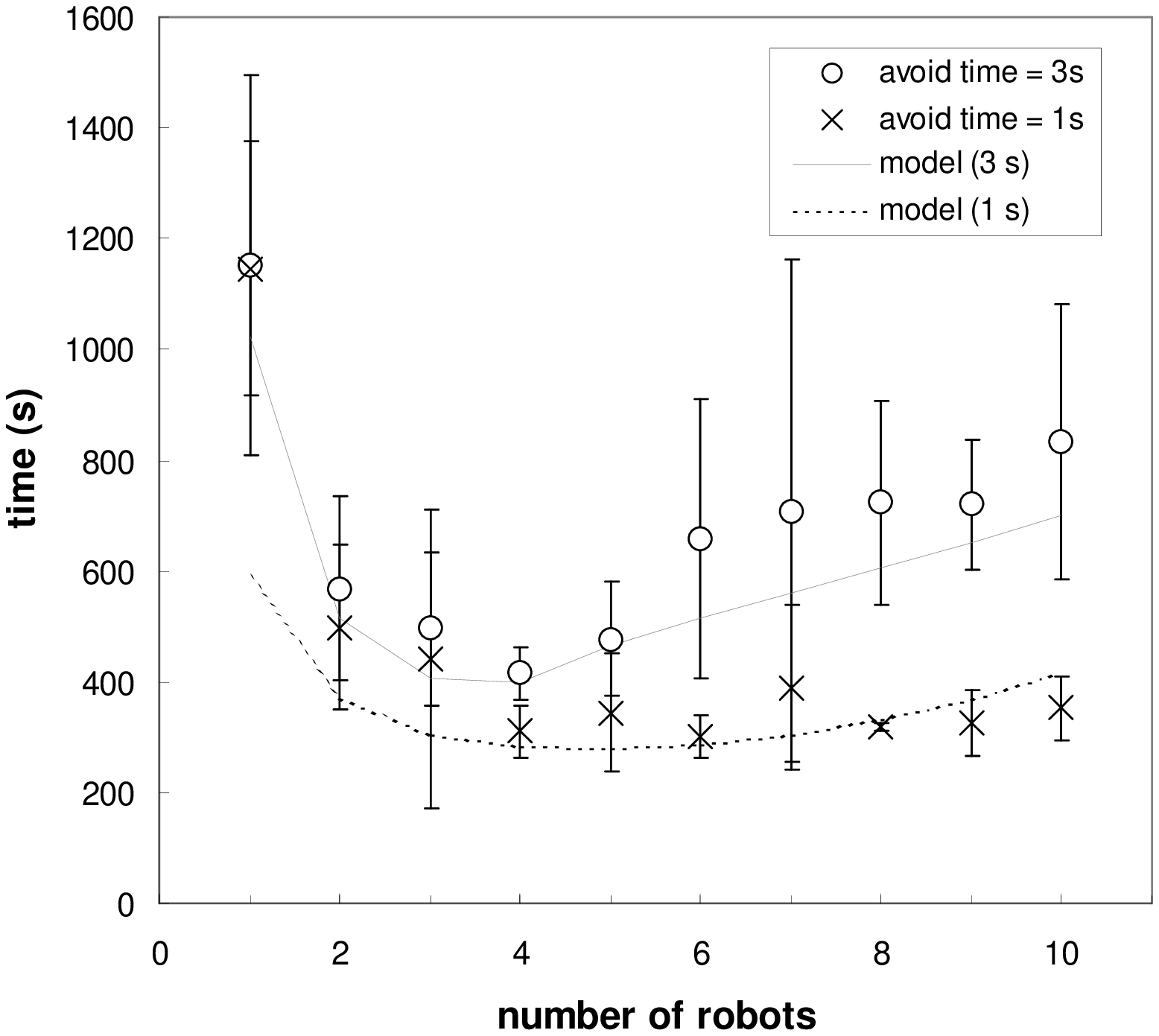}\\
  \epsfxsize = 2.6in
  \epsffile{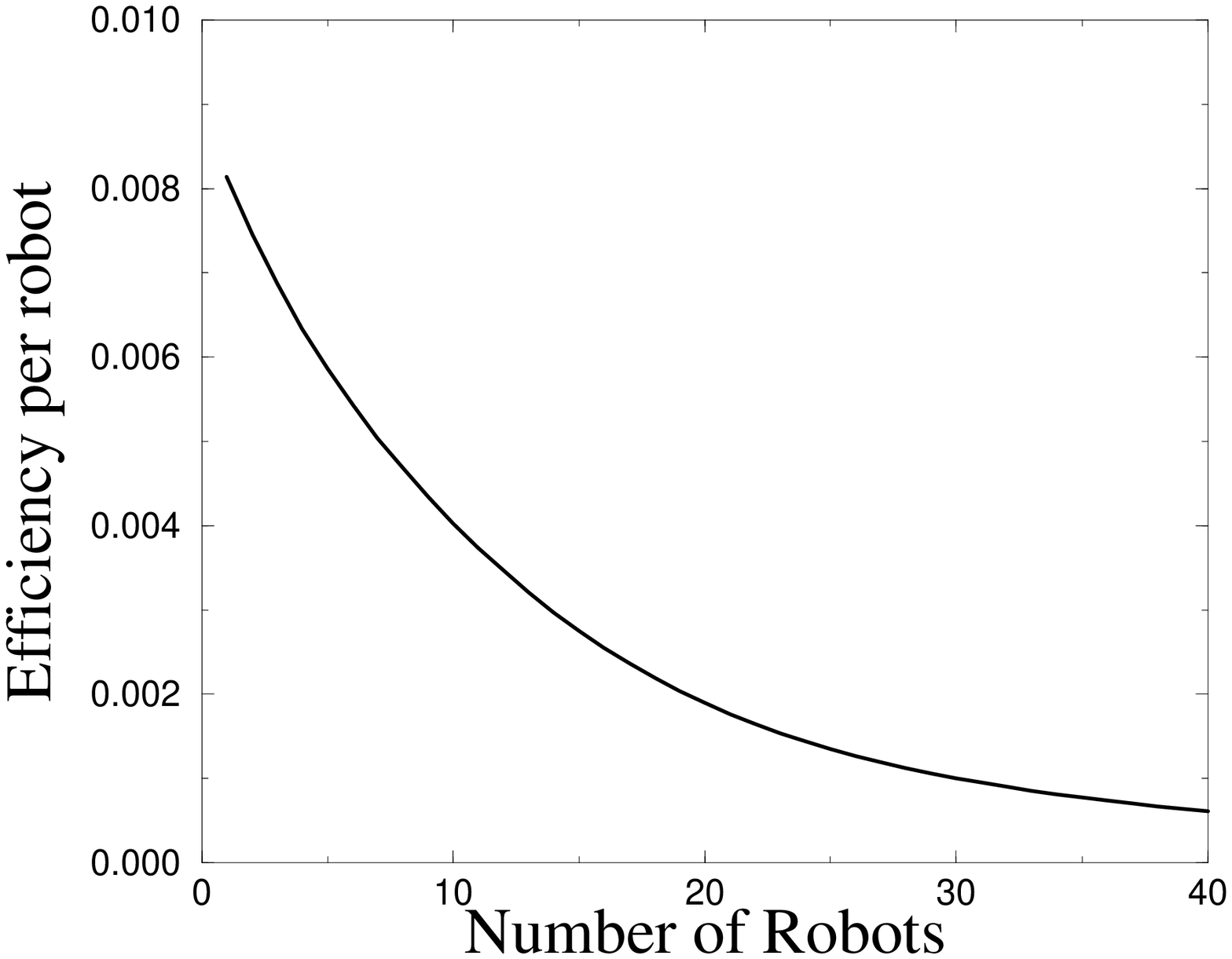}
  \end{tabular}
  \caption{The time it takes the group to collect and deliver pucks home for two interference strengths, $\tau = 3\ s$, and $\tau = 1.5 s$ and $\tau_h^0=16\ s$, $\alpha_p=0.015$, $\alpha_r = 0.04$, $\alpha_r^{\prime}=0.08$.
 (b) Efficiency per robot
  } \label{fig:homing-time}
  }
\end{figure}

Figure \ref{fig:homing-time}(a) shows the total time required to
complete the task for two different interference strengths, as
measured by the avoiding time parameter $\tau$. The solid line is
the result of the model's prediction for $\tau = 3\ s$, and the
dotted line for $\tau = 1.5\ s$\footnote{Although in the
simulations we specified the avoidance time to be 1\ s, multiple
collisions caused the average avoiding time per collision to be
slightly higher.}, and $\tau_h^0 = 16\ s$, $\alpha_p=0.015$,
$\alpha_r = 0.04$, and $\alpha_r^{\prime}=0.08$. \comment{We have
also taken into account the effect of wall avoidance, it's
strength given by $\alpha_w = 0.04$.} The simulations data shows
that the average avoiding time per collision increases as the
group size grows. This is due to multiple avoidance moves per each
attempt to avoid collision, caused by an increase in the local
density of robots. Therefore, in the model, we take the avoiding
time parameter $\tau$ as a linearly increasing function of $N_0$,
with the initial value of $\tau^0=3\ s$ (or $1.5\ s$). The
agreement between the model and simulations is good. \comment{Note
that there are minor differences between the values of the
parameters that best fit the data and the values we compute from
the simulations: we took $\tau_h^0 = 16.00\ s$ (the data shows the
average homing time for a single robot is $14.95 \pm 0.98$), and
$\alpha_p = 0.015$. Aside from these slight differences in the
parameter values, the agreement between model and simulations is
good. The differences are probably caused by the fact that our
model does not take into account the time robots spend in the
reverse homing behavior, during which they cannot search for
pucks.}

The final plot (Fig.~\ref{fig:homing-time}(b)) shows that
interference causes deterioration in relative performance. The per
robot efficiency is a monotonically decreasing function of the
group size. Thus, adding one new robot to the group decreases the
performance of every robot, though if the initial group size was
less than the optimal size, adding a robot will increase the
overall efficiency of the group.

\subsection{Foraging in Communicating Robots}
\label{sec:sugawara} 
\begin{figure}
  \epsfxsize 2.4in
  \center{
  \epsffile{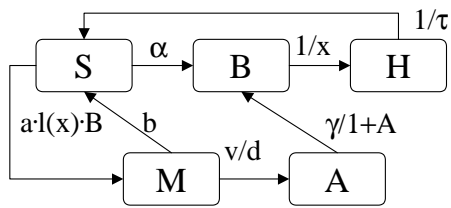}}
  \caption{State diagram of foraging robots from Sugawara {\em et al.}~\protect\cite{Sugawara97}}\label{fig:sugawara}
\end{figure}

Sugawara and
coworkers~\cite{Sugawara97,SugSanYosAbe98,Sugawara+al99} carried
out quantitative studies of foraging in groups of communicating
robots in different environments. In their system when a robot
finds a puck, it broadcasts a signal for a duration of time $x$.
If other robots detect the signal, they turn and move towards it.
After the interaction period, the broadcasting robot turns off the
signal, and makes a transition to the homing state. The
macroscopic state diagram for this system is shown in
Fig.~\ref{fig:sugawara} and is composed of the following states:
searching (S), broadcasting (B), homing (H), moving to the signal
source (M), and avoiding (A), with the corresponding dynamic
variables representing the average number of robots in each state.

\begin{eqnarray}\label{eq:sugawara}
  \frac {dN_h} {dt} & = & \frac{1}{(x+1)}N_b - \frac{1}{\tau} N_h \\
  \frac {dN_b} {dt} & = & - \frac{1}{(x+1)}N_b + \alpha N_s + \frac {\gamma} {a+N_a}N_a \\
  \frac {dN_s} {dt} & = & -\alpha N_s + nN_m + \frac{1}{\tau}N_h - a l(x) N_s N_b\\
  \frac {dN_a} {dt} & = & \frac{v}{d}N_m -\frac {\gamma} {a+N_a}N_a\\
  \frac {dN_m} {dt} & = & -\frac{v}{d}N_m -bN_m +a l(x) N_s N_b
\end{eqnarray}
\noindent where $\alpha$ is the probability to find a puck, $b$
probability to lose signal source, $\tau$ time to return home, $x$
interaction duration, $a$ is the probability of catching the
broadcast signal, $l(x)$ turn angle, $d$ average distance between
interacting robots, $v$ velocity of the robot, and $\gamma$
probability to find a puck by following other robots.

Interference between robots due to collision avoidance is assumed
to be negligible except during crowding near a broadcasting robot;
therefore, avoiding terms appear only in the equations describing
broadcasting and moving robots. The strength of interference is
phenomenologically described by a simple function that is
inversely proportional to the density of robots.

Sugawara \emph{et al.} found that for non-communicating robots
($x=0$), the time to complete the task was proportional to the
inverse of the number of robots --- $T \approx N^{-1}$. This is
true when robots work independently of one another. Interaction
through signal broadcast improves the efficiency of group behavior
--- $T \approx N^{\beta}$, $-1 < \beta < -2$  --- for most
durations $x$ of interaction. This result is independent of
whether the puck distribution was homogeneous or localized
(specified by parameter $\gamma$) . These findings were confirmed
by simulations and experiments with physical robots.

At first glance, the results of Sugawara \emph{et al.} appear to
contradict those presented in Fig.~\ref{fig:homing-time}(a), which
shows that performance time $T$ decreases with group size $N$ up
to some critical value and then starts to increase. \comment{In a
larger study of foraging in robots using the approach described
here~\cite{Lerman02a}, we showed that the time to complete a
collection task (foraging without homing) decreases monotonically
with $N$ as $T \approx N^{-0.7}$, even when avoiding behavior is
taken into account.} In fact, the results of both works support
the same basic conclusion that avoiding needs to be included in
the model when crowding conditions occur (the exact density
threshold remains to be determined). We believe the apparent
discrepancy in results can be explained by the difference in
systems being modeled. In Sugawara {\em et al.} work, ``home'' is
located at the center of the arena. Avoiding is not taken into
account in the no-interaction model because there is less crowding
near the home region than near a broadcasting robot. In the system
we studied, on the other hand, ``home'' is located at the edge of
the arena, and crowding is more pronounced. Both studies construct
a minimal model required to explain the observed behavior of the
system. The differences in the models and conclusions can be
traced back to the differences in the systems being modeled and
their behavior.

\section{Collaboration in Robots}
\label{sec:sp} Collaboration can significantly increase the
performance of a multi-agent system. In some systems collaboration
is an explicit requirement, because a single agent cannot
successfully complete the task on its own. Such ``strictly
collaborative''~\cite{Martinoli99} systems are common in insect
and human societies, {\em e.g.}, in transport of an object too
heavy or awkward to be lifted by a single ant, flying the space
shuttle, playing a soccer match, {\em etc}. Collaboration in a
group of robots has been studied by several
groups~\cite{MatNilSim95,KubBon00,MarMon95,IMB2001,KitTamStoVel98}.
We will focus on one group of experiments initiated by Martinoli
and collaborators~\cite{MarMon95} and studied by Ijspeert et
al.~\cite{IMB2001} that take a swarm approach to collaboration. In
this system collaboration in a homogeneous group of simple
reactive agents was achieved entirely through local interactions,
{\em i.e.}, without explicit communication or coordination among
the robots. Because they take a purely swarm approach, their
system is a compelling and effective model of how collaboration
may arise in natural systems, such as insect societies.

\subsection{Stick-pulling Experiments in Groups of Robots}
\begin{figure}[hptb]
\epsfxsize = 4.0in \center{\epsffile{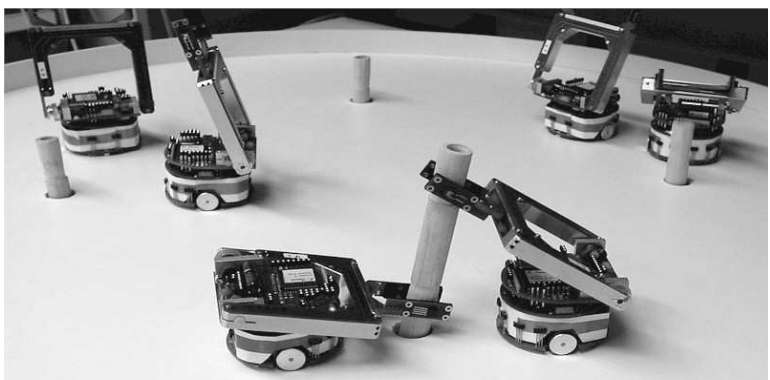}} 
\caption{Physical set-up of the stick-pulling experiment showing
six Khepera robots (courtesy of A.~Martinoli).}
\label{fig:experiment}
\end{figure}

The stick-pulling experiments were carried out by Ijspeert {\em et
al.} to investigate the dynamics of collaboration among locally
interacting simple reactive robots. Figure~\ref{fig:experiment} is
a snapshot of the physical set-up of the experiments. The robots'
task was to locate sticks scattered around the arena and pull them
out of their holes. A single robot cannot complete the task (pull
the stick out) on its own --- a collaboration between two robots
is necessary for the task to be successfully completed. In a more
general case, a collaboration between an arbitrary number of
robots may be required to successfully complete the tasks (because
sticks may be of varying length). The collaboration occurs in the
following way: one robot finds a stick, and waits for a second
robot to find it, lifting the stick partially out of its hole.
When a second robot finds it, it will grip the stick and pull it
out of the ground, successfully completing the task. (In the
general case, a group of some size has to accumulate at the site
of the stick before the required number of robots necessary to
complete the task is present.)

\begin{figure}[hptb]
\epsfxsize = 3.0in \center{\epsffile{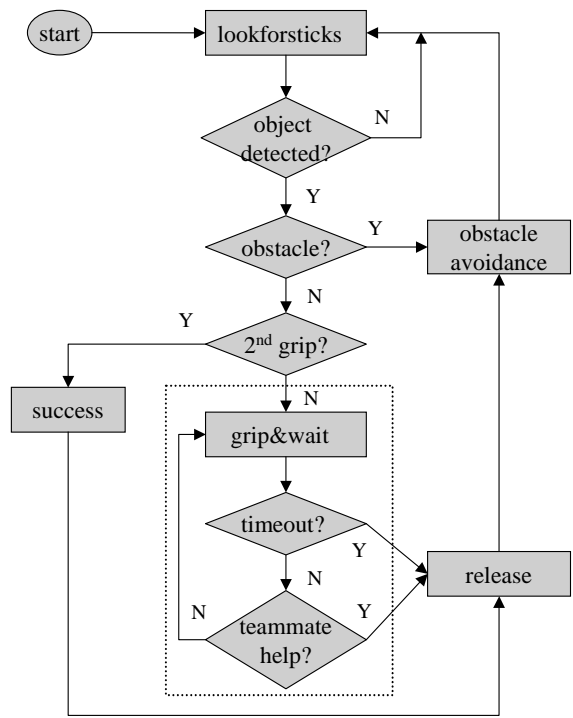}}
\caption{Flowchart of the robots' controller (from Ijspeert et
al~\protect\cite{IMB2001}). } \label{fig:controller}
\end{figure}

The actions of each robot are governed by a simple controller,
outlined in Figure~\ref{fig:controller}. The robot's default
behavior is to wander around the arena looking for sticks and
avoiding obstacles, which could be other robots or walls. When a
robot finds a stick that is not being held by another robot, it
grips it, lifts it half way out of the ground and waits for a
period of time specified by the {\em gripping time parameter}. If
no other robot comes to its aid during the waiting period, the
robot releases the stick and resumes the search for other sticks.
If another robot encounters a robot holding a stick, a successful
collaboration will take place during which the second robot will
grip the stick, pulling it out of the ground completely, while the
first robot releases the stick and resumes the search. After the
task is completed, the second robot also releases the stick and
returns to the search mode, and the experimenter replaces the
stick in its hole.

Ijspeert {\em et al.} studied the dynamics of collaboration in
stick-pulling robots on three different levels: by conducting
experiments with physical robots; with a sensor-based simulator of
robots; and using a probabilistic microscopic model. The physical
experiments were performed with groups of two to six Khepera
robots in an arena containing four sticks. Because experiments
with physical robots are very time consuming, Webots, the
sensor-based simulator of Khepera robots~\cite{Michel98}, was used
to systematically explore the parameters affecting the dynamics of
collaboration. Webots simulator attempts to faithfully replicate
the physical experiment by reproducing the robots' (noisy) sensory
input and the (noisy) response of the on-board actuators in order
to compute the trajectory and interactions of each robot in the
arena. The probabilistic microscopic model, on the other hand,
does not attempt to compute the trajectories of individual robots.
Rather, the robot's actions
--- encountering a stick, a wall, another robot, a robot gripping
a stick, or wandering around the arena --- are represented as a
series of stochastic events, with probabilities based on simple
geometric considerations. For example, the probability of a robot
encountering a stick is equal to the product of the number of
ungripped sticks, and the detection area of the stick normalized
by the arena area. Probabilities of other interactions can be
similarly calculated. The microscopic simulation consists of
running several processes in parallel, each representing a single
robot, while keeping track of the global state of the environment,
such as the number of gripped and ungripped sticks. According to
Ijspeert {\em et al.} the acceleration factor for Webots and real
robots can vary between one and two orders of magnitude for the
experiments presented here. Because the probabilistic model does
not require calculations of the details of the robots'
trajectories, it is $300$ times faster than Webots for these
experiments.

\subsubsection{Experimental Results}
\label{sec:martinoliresults} 
\begin{figure}[hptb]
\epsfxsize = 3.0in \center{\epsffile{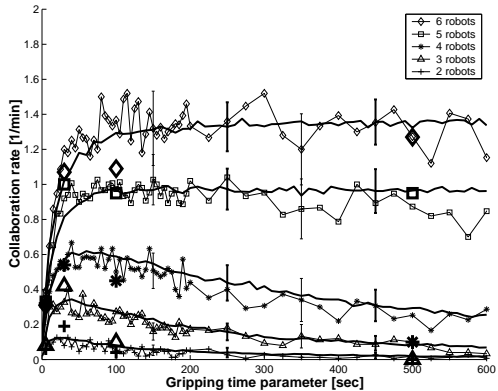}}
\caption{Collaboration rate vs. the gripping time parameter for
groups of two to six robots and four sticks (from Ijspeert {\em et
al}). Heavy symbols represent experimental results, symbols
connected by lines are the results of sensor-based simulations,
while the smooth heavy lines are the results of the probabilistic
microscopic model.} \label{fig:martinoliresults}
\end{figure}

Ijspeert {\em et al.} systematically
studied the collaboration rate (the number of sticks successfully
pulled out of the ground in a given time interval), and its
dependence on the group size and the gripping time parameter. They
found very good qualitative and quantitative agreement between the
three different levels of experiments, as shown in
Figure~\ref{fig:martinoliresults}. Their main observation was
that, depending on the ratio of robots to sticks (or workers to
the amount of work), there appear to be two different regimes in
the collaboration dynamics. When there are fewer robots than
sticks, the collaboration rate decreases to zero as the value of
the gripping time parameter grows. In the extreme case, when the
robot grabs a stick and waits indefinitely for another robot to
come and help it, the collaboration rate is zero, because after
some period of time each robot ends up holding a stick, and no
robots are available to help. When there are more robots than
sticks, the collaboration rate remains finite even in the limit
the gripping time parameter becomes infinite, because there will
always be robots available to help pull the sticks out. Another
finding of Ijspeert {\em et al.} was that when there are fewer
robots than sticks, there is an optimal value of the gripping time
parameter which maximizes the collaboration rate. In the other
regime, the collaboration rate appears to be independent of the
gripping time parameter above a specific value, so the optimal
strategy is for the robot to grip a stick and hold it
indefinitely.

In the following section we present a macroscopic mathematical
model of the stick-pulling experiments in a homogeneous
multi-robot system. Such a model is useful for the following
reasons. First, the model is independent of the system size, {\em
i.e.} the number of robots; therefore, solutions for a system of
$5,000$ robots take just as long to obtain as solutions for $5$
robots, whereas for a microscopic description the time required
for simulation scales at least linearly with the number of robots.
Second, our approach allows us to directly estimate certain
important parameter values, ({\em e.g.}, those for which the
performance is optimal) without having to resort to the time
consuming simulations or experiments. It also enables us to study
the stability properties of the system, and see whether solutions
are robust under external perturbation or noise. These
capabilities are important for the design and control of large
multi-agent systems.

\subsection{The Rate Equation Model of Collaboration}
In order to construct a mathematical model of collective behavior
in stick-pulling experiments, it is helpful to draw the
macroscopic state diagram of the system. On a macroscopic level,
during a sufficiently short time interval, each robot will be in
one of two states: $searching$ or $gripping$. Using the flowchart
of the robots' controller (Fig.~\ref{fig:controller}) as
reference, we include in the search state the set of behaviors
associated with the looking for sticks mode, such as wandering
around the arena (``look for sticks'' action), detecting objects
and avoiding obstacles; while the gripping state is composed of
decisions and an action inside the dotted box. We assume that
actions ``success'' (pull the stick out completely) and
``release'' (release the stick) take place on a short enough time
scale that they can be incorporated into the search state. Of
course, there can be a discrete state corresponding to every
action depicted in Fig.~\ref{fig:controller}, but this would
complicate the mathematical analysis without adding much to the
descriptive power of the model. While the robot is in the obstacle
avoidance mode, it cannot detect and try to grip objects;
therefore, avoidance serves to decrease the number of robots that
are searching and capable of gripping sticks. We studied the
effect of avoidance in \cite{Lerman01a} and found that it does not
qualitatively change the results of the simpler model that does
not include avoidance; therefore, we will leave it out for
clarity.

\begin{figure}[hptb]
\epsfxsize = 2.0in \center{\epsffile{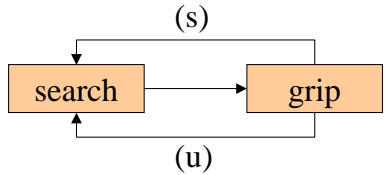}}
\caption{Macroscopic state diagram of the multi-robot system. The
arrow marked 's' corresponds to the transition from the gripping
to the searching state after a successful collaboration, while the
arrow marked 'u' corresponds to the transition after an
unsuccessful collaboration, {\em i.e.}, when the robots releases
the stick without a successful collaboration taking place. }
\label{fig:macro}
\end{figure}

In addition to states, we must also specify all possible
transitions between states. When it finds a stick, the robot makes
a transition from the search state to the gripping state. After
both a successful collaboration and when it times out
(unsuccessful collaboration) the robot releases the stick and
makes a transition into the searching state, as shown in
Fig.~\ref{fig:macro}. These arrows correspond to the arrow
entering and the two arrows leaving the dotted box in
Fig.~\ref{fig:controller}. We will use the macroscopic state
diagram as the basis for writing down the rate equations that
describe the dynamics of the stick-pulling experiments. Note that
the experimental system was a semi-Markov system, because the
robot's transition from gripping to the searching state depended
not only on its present state (gripping) but also on how long the
robot has been in the gripping state.

We can simplify analysis by considering a modified version of the
robot controller presented in Fig.~\ref{fig:controller}, where
instead of waiting a specified period of time, each robot releases
the stick with some probability per unit time. Note that this
simplification restores the Markovian property of the system. We
also construct and analyze a more complex model
(Sec.~\ref{sec:gripping}) that explicitly includes the gripping
time parameter. We show that the system based on the simplified
controller produces qualitatively the same macroscopic behavior as
the original system, which is modeled by the second and more
complex model. Both systems are described by the same macroscopic
state diagram and differ only in the details of the transition
rate between the gripping and searching states.

Each box in Fig.~\ref{fig:macro} corresponds to a macroscopic
state and therefore requires a dynamic variable to describe it.
Thus, the variables of our model are $N_s(t)$ and $N_g(t)$, the
number of robots in the searching and gripping states
respectively. Also, let  $M(t)$ be the number of uncollected
sticks  at time $t$. The latter variable does not represent a
macroscopic state, rather it tracks the state of the environment.
The mathematical model of the stick-pulling experiments consists
of a series of coupled rate equations, each describing how the
dynamic variables evolve in time:
\begin{eqnarray}
\label{eq:ode1} \frac{dN_s}{dt}& =& -\alpha N_s(t)\biggl (
M(t)-N_g(t)\biggr ) + \tilde{\alpha} N_s(t)N_g(t) +\gamma N_g(t) \,,\\
\label{eq:odeN}
N_0 & = & N_s + N_g  \,,\\
\label{eq:odeM} \frac{dM}{dt}& = & -\tilde{\alpha}N_s(t)N_g(t) +
\mu(t) \,,
\end{eqnarray}
where $\alpha$, $\tilde{\alpha}$ are the rates at which a
searching robot encounters a stick and a gripping robot
respectively, $\gamma$ is the rate at which robots release sticks
and $\mu(t)$ is the rate at which new tasks are added. The
parameters $\alpha$, $\tilde{\alpha}$, and $\gamma$ connect the
model to the experiment: $\alpha$ and $\tilde{\alpha}$ are related
to the size of the object, the robot's detection radius, or
footprint, and the speed at which it explores the arena.

The three terms in Eq.~\ref{eq:ode1} correspond to the three
arrows between the states in Fig.~\ref{fig:macro}. The first term
accounts for the decrease in the number of searching robots
because some robots find and grip sticks; the second term
describes the successful collaborations between two robots, and
the third term accounts for the failed collaborations (\ie, when a
robot releases a stick without a second robot present), both of
which lead to an increase the number of searching robots. We do
not need a separate equation for $N_g$, since this quantity may be
calculated from the conservation of robots condition,
Eq.~\ref{eq:odeN}. The last equation states that the number of
sticks, $M(t)$, decreases in time at the rate of successful
collaborations, Eq.~\ref{eq:odeM}. The equations are subject to
the initial conditions that at $t=0$ the number of searching
robots in $N_0$ and the number of sticks is $M_0$.

To proceed further, let us introduce
\begin{eqnarray}
\label{eq:sp-defs}
n(t)&=&N_s(t)/N_0,\\
m(t)&=&M(t)/M_0, \\
\beta&=&N_0/M_0, \\
R_G&=&\tilde{\alpha}/\alpha,\\
\label{eq:sp-defsN} \tilde{\beta}&=&R_G \beta
\end{eqnarray}
and dimensionless time $t \rightarrow \alpha M_0 t$. Here $n(t)$
is the fraction of robots in the search state and $m(t)$ is the
fraction of uncollected sticks at time $t$. Due to the
conservation of number of  robots, the fraction of robots in the
gripping state is simply $1 - n(t)$. Equations
\ref{eq:ode1}--\ref{eq:odeM} can be rewritten in dimensionless
form as:
\begin{eqnarray}
\label{eq:ode2} \frac{dn}{dt}& = & -n(t)[m(t)+\beta
n(t)-\beta]+\tilde{\beta}n(t)[1-n(t)] + \gamma [1-n(t)] \\
\label{eq:ode3} \frac{dm}{dt}& = & -\beta
\tilde{\beta}n(t)[1-n(t)] +\mu^{\prime}
\end{eqnarray}
Equations \ref{eq:ode2}--\ref{eq:ode3} together with initial
conditions $n(0)=1$, $m(0)=1$ determine the dynamical evolution of
the system. Note that only two parameters, $\beta$ and $\gamma$,
appear in the equations and, thus, determine the behavior of
solutions. The third parameter $\tilde{\beta}=R_G \beta$ is fixed
experimentally and is not independent. Note that we do not need to
specify $\alpha$ and $\tilde{\alpha}$ --- they enter the model
only through $R_G$ (throughout this paper we will use
$R_G=0.35$).\footnote{The parameter $\alpha$ can be easily
calculated from experimental values quoted in
\protect\cite{IMB2001}. As a robot travels through the arena, it
sweeps out some area during time $dt$ and will detect objects that
fall in that area. This detection area is $V_R W_R dt$, where
$V_R=8.0\ cm/s$ is robot's speed, and $W_R=14.0\ cm$ is robot's
detection width. If the arena radius is $R=40.0\ cm$, a robot will
detect sticks at the rate $\alpha = V_R W_R /\pi R^2  = 0.02\
s^{-1}$. According to \protect\cite{IMB2001}, a robot's
probability to grab a stick already being held by another robot is
$35\%$ of the probability of grabbing a free stick. Therefore,
$R_G = \tilde{\alpha}/\alpha = 0.35$. $R_G$ is an experimental
value obtained  with systematic experiments with two real robots,
one holding the stick and the other one approaching the stick from
different angles. }

We assume that new sticks are added to the system at the rate that
the robots pull them out; therefore, the number of sticks does not
change with time ($m(t)=m(0)=1$). This  situation may be realized
experimentally by replacing the sticks in their holes after they
are pulled out by robots.  A steady-state solution, if it exists,
describes the long term time-independent behavior of the system.
To find it, we set the left hand side of Eq.~\ref{eq:ode2} to
zero:
\begin{equation}
\label{eq:steady-state} -n[1+\beta n-\beta]+\tilde{\beta}n[1-n] +
\gamma [1-n]=0.
\end{equation}
\noindent This quadratic equation can be solved to obtain steady
state values of $n(\beta,\gamma)$.

\begin{figure}[tp]
\begin{tabular}{cc}
\epsfxsize = 2.1in {\epsffile{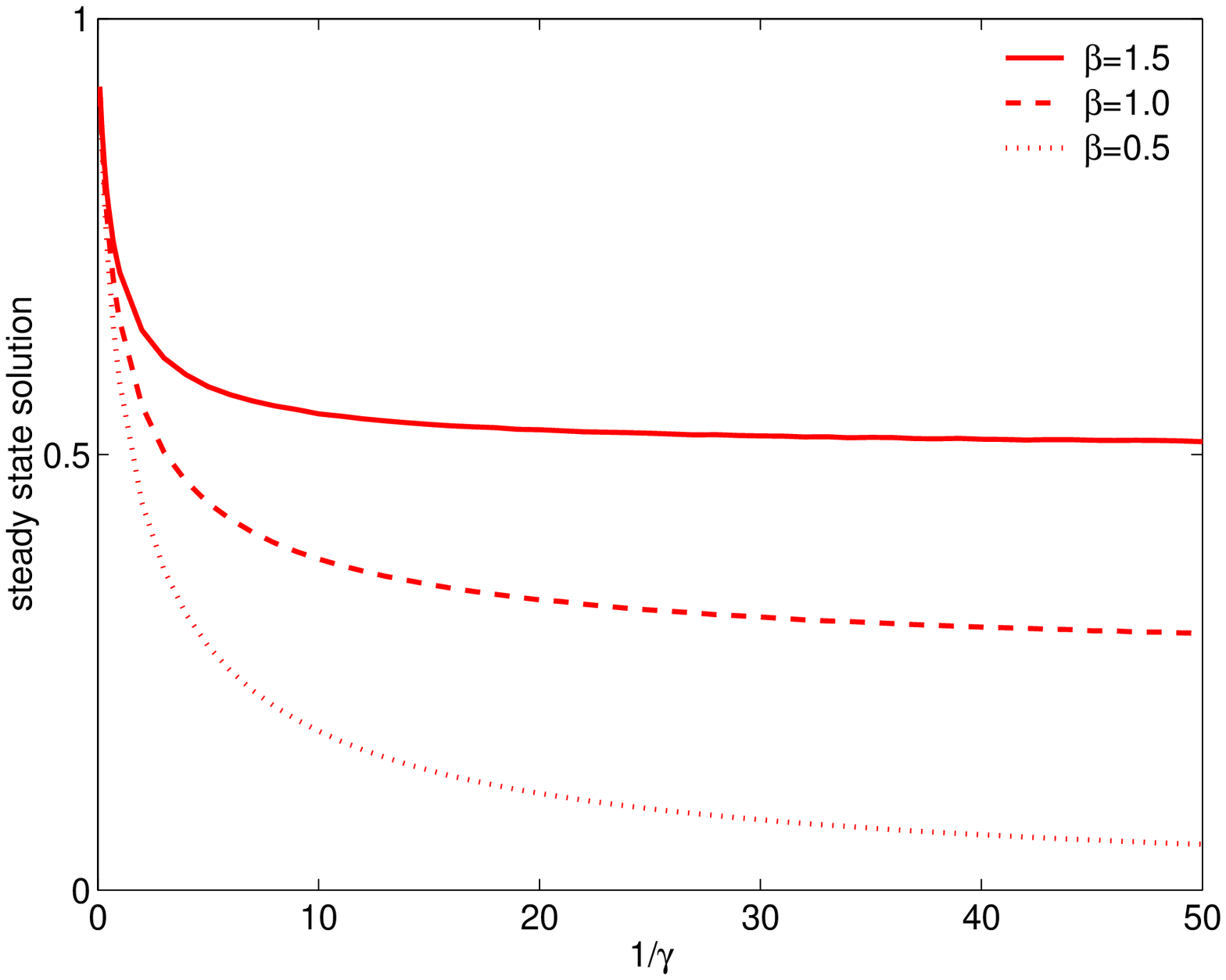}} & \epsfxsize =
2.1in {\epsffile{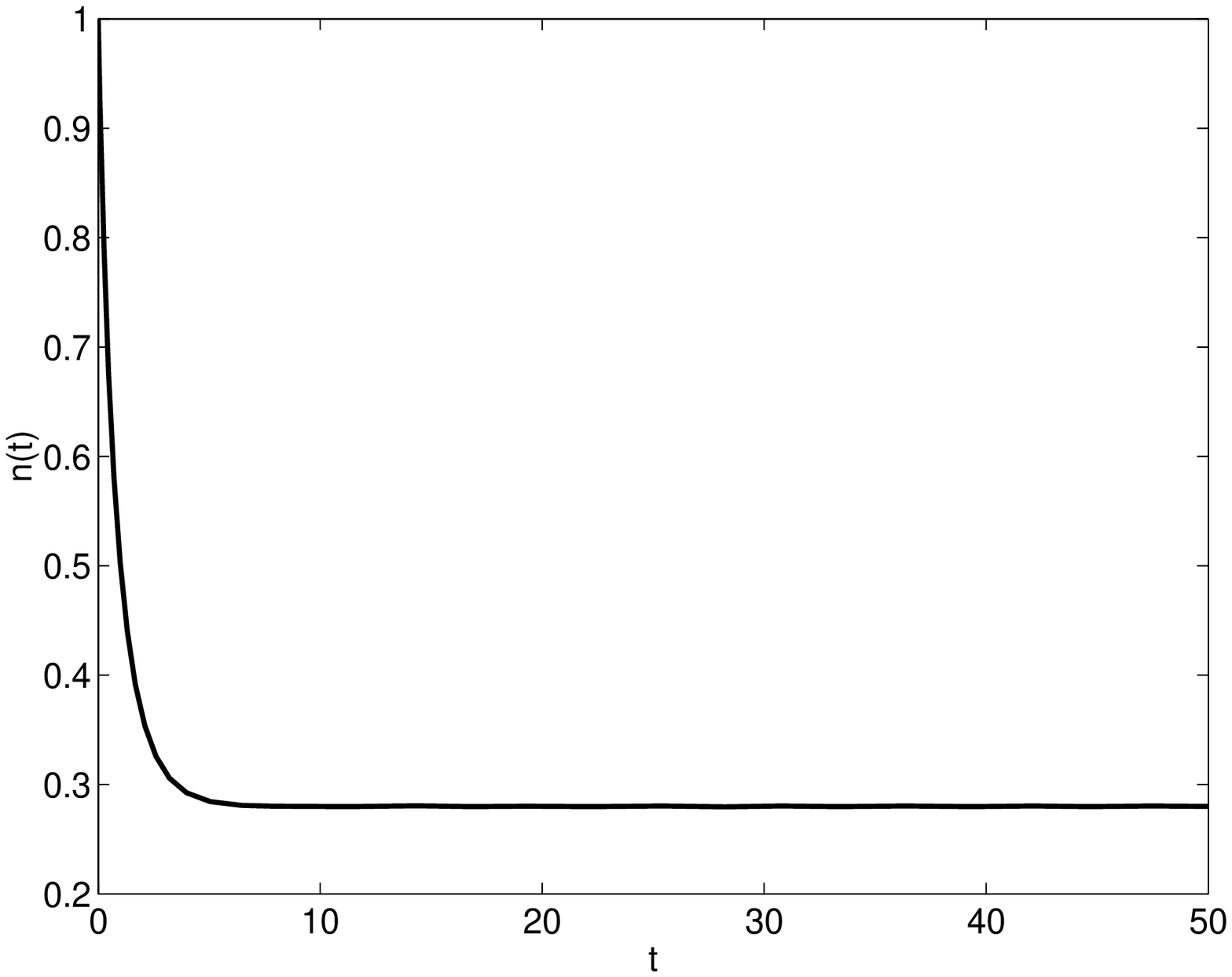}}\\
(a)&(b)\\
\end{tabular}
\caption{ (a) Steady state solution vs inverse stick release rate
$1/\gamma$. (b) Typical relaxation  to the steady state of the
fraction of searching robots for $\gamma=0.2$, $\beta=0.5$.}
\label{fig:sp_steady_state}
\end{figure}

Figure \ref{fig:sp_steady_state}(a) shows the dependence of the
fraction of searching robots in the steady state on the parameters
$\beta$ and $\gamma$. The x-axis has units of time, although the
scale is different than in Fig.~\ref{fig:martinoliresults}. Note,
that for small enough $\beta$'s $n(\gamma)\rightarrow 0$ as
$1/\gamma \rightarrow \infty$. The intuitive reason for this is
the same one given in Section~\ref{sec:martinoliresults}: when
there are fewer robots than sticks, and each robot holds the stick
indefinitely (vanishing release probability), after a while every
robot is holding a stick, and no
robots are searching. For larger values of $\beta$, 
however, $n(\gamma)\rightarrow const \neq 0$ as $1/\gamma
\rightarrow \infty$. Figure~\ref{fig:sp_steady_state}(b) shows how
a typical solution $n(t)$ relaxes to its steady state value.

\begin{figure}[thbp]
\epsfxsize = 3.2in \center{\epsffile{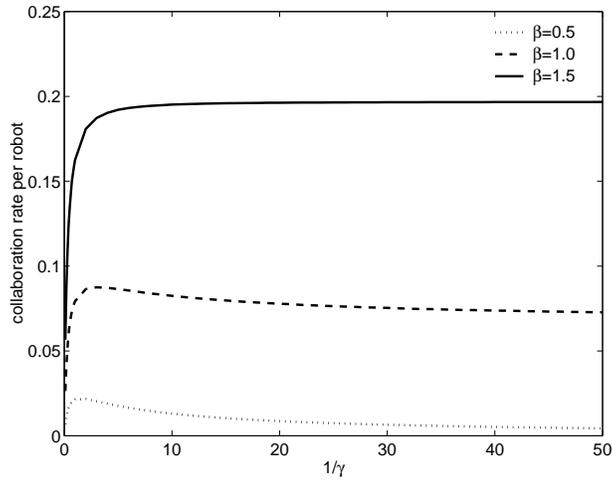}} \vskip 0.1in
\caption{ Collaboration rate per robot vs inverse stick release
rate $1/\gamma$ for $\beta=0.5$, $\beta=1.0$, $\beta=1.5$. These
values of $\beta$ correspond, respectively, to two, four, and six
robots in the experiments with four sticks ({\em cf.}
Fig.~\protect{\ref{fig:martinoliresults}}).}
\label{fig:sp_collaboration}
\end{figure}

The collaboration rate is the rate at which robots successfully
pull sticks out of their holes. The steady-state collaboration
rate $R(\gamma;\beta)$ is given by the following equation:
\begin{equation}
\label{eq:collaboration} R(\gamma,\beta)=\beta \tilde{\beta}
n(\gamma,\beta)[1-n(\gamma,\beta)]\,,
\end{equation}
\noindent where $n(\gamma,\beta)$ is the steady-state number of
searching robots for a particular value of $\gamma$ and $\beta$,
and $(1 - n(\gamma, \beta))$ is the steady-state number of
gripping robots.  Figure \ref{fig:sp_collaboration} depicts the
collaboration rate as a function of $1/\gamma$. For $\beta >
\beta_c$ the collaboration rate increases monotonically with
$1/\gamma$. However, for
 $\beta < \beta_c$ there is an optimal stick release rate which
maximizes the collaboration rate. The optimal value of $\gamma$
which maximizes the collaboration rate can be computed from the
condition $dR(\gamma, \beta)/d\gamma=\beta \tilde{\beta}
d(n-n^2)/d \gamma=0$, with $n$ given by roots of
Eq.~\ref{eq:steady-state}. Another way to compute the optimal
release rate is by noting that for a given value of $\beta$ below
some critical value, the collaboration rate is greatest when half
of the robots are gripping and the other half are searching.
Substituting $n=1/2$ into Eq.~\ref{eq:steady-state}, leads to
$\gamma_{opt}=1-(\beta+\tilde{\beta})/2$. $\gamma_{opt}$ vanishes
as $\beta$ exceeds its critical value, $\beta_c=2/(1+R_G)$;
therefore, for $\beta >\beta_c$, $n>1/2$, and no optimal release
rate exists.

The three curves in Fig.~\ref{fig:sp_collaboration} are
qualitatively similar to those in Fig.~\ref{fig:martinoliresults}
for $2$ robots ($\beta = 0.5$), $4$ robots ($\beta = 1.0$) and $6$
robots ($\beta = 1.5$). Even the grossly simplified model
reproduces the following conclusions of Ijspeert {\em et al.}: the
different dynamical regimes depending on the value of the ratio of
robots to sticks ($\beta$) and the optimal gripping time parameter
for $\beta < \beta_c$.

In the next section, we construct a model of the stick pulling
experiments that explicitly includes the gripping time parameter
$\tau$. The collective behavior predicted by the more accurate
model is both qualitatively and quantitatively similar to that
predicted by the simplified model.

\subsection{Model with Gripping Time Parameter}
\label{sec:gripping} A more accurate mathematical description of
the stick pulling experiments should explicitly includes the
gripping time parameter $\tau$. Note that the system is now a
semi-Markov system, because transitions from gripping to the
searching state depend not only on the present state (gripping)
but also on how long the robot has been in the gripping state,
{\em i.e.}, whether or not it has timed out. This property of the
system will be captured by time-dependent transition rates. The
system is described by the same macroscopic state diagram,
Fig.~\ref{fig:macro}, and the same set of equations,
Eq.~\ref{eq:ode1}--\ref{eq:odeM}, with only the last term in
Eq.~\ref{eq:ode1} different. We write the new equation below:
\begin{eqnarray}
\label{eq:ode1a}
\frac{dN_s}{dt}& =& -\alpha N_s(t)\biggl ( M(t)-N_g(t)\biggr ) + \tilde{\alpha} N_s(t)N_g(t) \nonumber \\
& & +\alpha N_s(t-\tau)\biggl ( M(t-\tau)-N_g(t-\tau)\biggr )
\Gamma(t;\tau).
\end{eqnarray}
All the parameters have the same meaning as before.
$\Gamma(t;\tau)$, the fraction of failed collaborations at time
$t$, is the probability no robot came ``to help'' during the time
interval $[t-\tau,t]$. This is a time-dependent parameter, and it
describes unsuccessful transitions from the gripping state in this
semi-Markov system. To calculate $\Gamma(t;\tau)$ let us divide
the time interval $[t-\tau,t]$ into $K$ small intervals of length
$\delta t=\tau/K$. The probability that no robot comes to help
during the time interval $[t-\tau,t-\tau+\delta t]$ is simply
$1-\tilde{\alpha} N_s(t-\tau)\delta t$. Hence, the probability for
a failed collaboration is
\begin{eqnarray}
\Gamma(t;\tau)& =& \prod_{i=1}^{K}[1-\tilde{\alpha}\delta t N_s(t-\tau+i\delta t )] \Theta (t-\tau) \nonumber \\
\label{fc1} & \equiv & \exp\biggl [\sum_{i=1}^{K}
\ln[1-\tilde{\alpha}\delta t N_s(t-\tau+i \delta t)]\biggr ]
\Theta (t-\tau)
\end{eqnarray}
The step function $\Theta(t-\tau)$ ensures that $\Gamma (t; \tau)$
is zero for $t<\tau$. Finally, expanding the logarithm in
Eq.~(\ref{fc1}) and taking the limit $\delta t \rightarrow 0$ we
obtain
\begin{equation}
\label{fc2}
\Gamma(t;\tau)=\exp[-\tilde{\alpha}\int_{t-\tau}^{t}dt^{\prime}N_s(t^{\prime})]
\Theta(t - \tau)
\end{equation}

We rewrite the model in dimensionless form using variable
transformations given by Eqs.~\ref{eq:sp-defs}--\ref{eq:sp-defsN}
and dimensionless gripping time parameter $\tau \rightarrow \alpha
M_0 \tau$:
\begin{eqnarray}
\label{eq:ode2a} \frac{dn}{dt}& = & -n(t)[m(t)+\beta
n(t)-\beta]+\tilde{\beta}n(t)[1-n(t)]\nonumber \\
& & + n(t-\tau)[m(t-\tau)+\beta n(t-\tau)-\beta]\times \gamma(t;\tau) \\
\label{eq:ode3a}
\frac{dm}{dt}& = & -\beta \tilde{\beta}n(t)[1-n(t)] +\mu^{\prime}\\
\label{eq:gamma} \gamma(t;\tau)& = &
\exp[-\tilde{\beta}\int_{t-\tau}^{t}dt^{\prime}n(t^{\prime})]
\end{eqnarray}

Equations \ref{eq:ode2a}--\ref{eq:gamma} are solved subject to
initial conditions $n(0)=1$ and $m(0)=1$ to determine the dynamic
evolution of the system. Parameters $\beta$ and $\tau$ alone
appear in the equations and thus, determine the behavior of the
system.

\begin{figure}[tp]
\epsfxsize = 3.2in \center{\epsffile{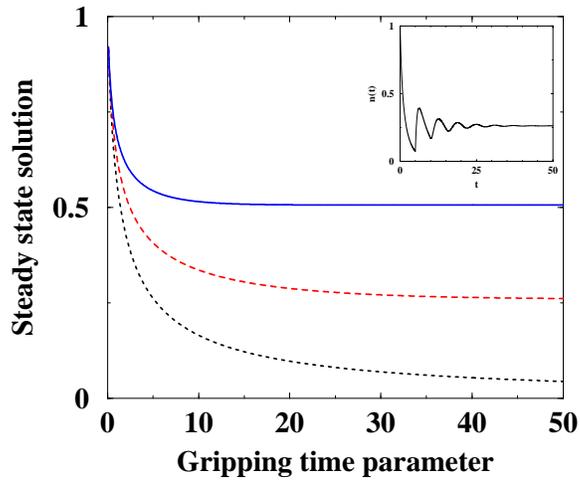}} \vskip 0.1in
\caption{ Steady state solution vs (dimensionless) gripping time
parameter $\tau$: for $\beta=0.5$ (short dash), $1$ (long dash),
$1.5$ (solid line). Inset shows a typical relaxation to the steady
state for $\tau=5$, $\beta=0.5$.}
\label{fig:sp_static_steady_state}
\end{figure}

Equation \ref{eq:ode2a} has a non-trivial steady-state solutions
which satisfy the following transcendental equation:
\begin{equation}
\label{trans}
-1+(\beta+\tilde{\beta})(1-n)+(1-\beta(1-n))e^{-\tilde{\beta}\tau
n}=0
\end{equation}
Figure \ref{fig:sp_static_steady_state} shows the dependence of
the steady state fraction of searching robots on the gripping time
$\tau$ for different values of $\beta$. Note, that for small
enough $\beta$'s $n(\tau)\rightarrow 0$ as $\tau \rightarrow
\infty$, {\em i.e.}, when there are fewer robots than sticks, and
each robot holds the stick indefinitely, after a while every robot
is holding a stick, and no robots are searching. For $\beta >
1/(1+R_G)$, however, $n(\tau)\rightarrow const \neq 0$ as $\tau
\rightarrow \infty$. The inset in
Fig.~\ref{fig:sp_static_steady_state} shows how a typical
solution, $n(t)$, relaxes to its steady state value. The
oscillations are characteristic of time-delay differential
equations, and their period is determined by $\tau$.

\begin{figure}[bp]
\epsfxsize = 3.2in \center{\epsffile{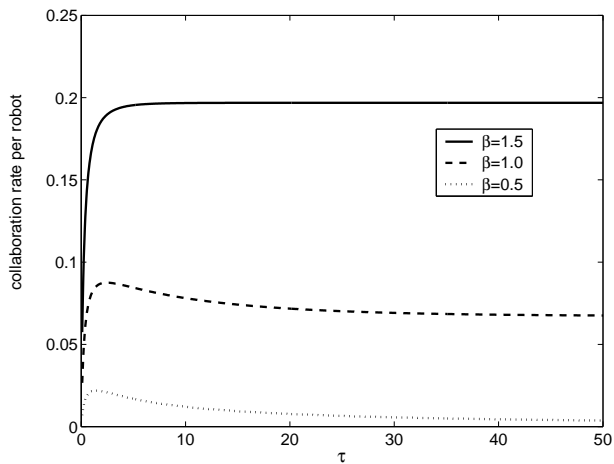}} \vskip 0.1in
\caption{ Collaboration rate per robot vs (dimensionless) gripping
time parameter $\tau$ for $\beta=0.5$ (short dash), $\beta=1$
(long dash), $\beta=1.5$ (solid line). These values of $\beta$
correspond, respectively, to two, four, and six robots in the
experiments with four sticks ({\em cf.}
Fig.~\protect{\ref{fig:martinoliresults}}).}
\label{fig:sp_static_collaboration}
\end{figure}

\begin{figure}[tp]
\epsfxsize = 2.0in \center{\epsffile{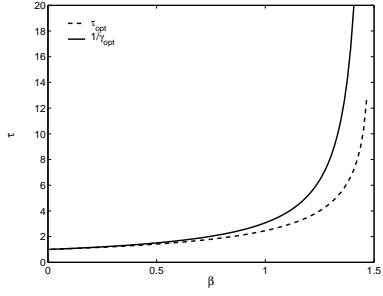}} \vskip 0.1in
\caption{ Optimal gripping time parameter and inverse of the
optimal release rate vs $\beta$ in the two models}
\label{fig:optimal}
\end{figure}

The steady--state collaboration rate $R(\tau;\beta)$, the rate at
which robots pull sticks out of their holes, is given by:
$R(\tau,\beta)=\beta \tilde{\beta} n(\tau,\beta)[1-n(\tau,\beta)]$
where $n(\tau,\beta)$ is the number of searching robots in the
steady-state for a particular value of $\tau$ and $\beta$ (and $(1
- n(\tau, \beta))$ is the number of gripping robots in the
steady-state). Figure \ref{fig:sp_static_collaboration} depicts
collaboration rate as a function of $\tau$. For $\beta > \beta_c$
the collaboration rate increases monotonically with $\tau$.
However, for $\beta < \beta_c$ there is an optimal gripping time,
$\tau=\tau_{opt}$, which maximizes the collaboration rate. We use
the same arguments as before to understand this behavior: maximum
collaboration rate for a given $\beta$ is achieved for
$n(\tau,\beta)=1/2$. For $\beta > \beta_c$, however, the solution
of  Eq.~\ref{trans} is always greater than $1/2$, so an optimal
solution does not exist. For $\beta < \beta_c$, simple analysis
gives
\begin{equation}
\tau_{opt}=\frac{2}{\tilde{\beta}}\ln\frac{1-
\beta/2}{1-1/2(\beta+\tilde{\beta})}, \ \ \beta <
\beta_c=\frac{2}{1+R_G}
\end{equation}
\noindent This dependence on $R_G$ was quantitatively confirmed
through embodied and microscopic
simulations~\cite{MartinoliEaston02}. Figure \ref{fig:optimal}
compares the optimal gripping time parameter and inverse of the
optimal release rate predicted by the two models. For $\beta < 1$
the two models give $quantitatively$ similar results, in addition
to predicting the same $\beta_c$. This example illustrates our
claim that in many cases a minimal model is sufficient to explain
and predict interesting system properties. Ref. \cite{Lerman01a}
presents more details of the analysis of the collaboration task,
including the effects of interference.

\subsection{Difference Equation Model of Collaboration}
Martinoli and Easton~\cite{MartinoliISER02,MartinoliEaston02}
consider a more fine-grained macroscopic model than the one
described above that takes into account more of the individual
robot behaviors shown in Fig.~\ref{fig:controller}. Their model
consists of coupled finite difference equations, one for each
state: searching ($N_s$), avoiding ($N_a$), interference ($N_i$),
success dance ($N_d$), and gripping ($N_g$). The equation for how
the number for searching robots changes in time is presented
below; equations for other variables are similar.
\begin{eqnarray*}\label{eqn:collaboration-difference}
  N_S(k+1) & = & N_s(k) - (\alpha_w+\alpha_r)N_s(k)
  -\tilde{\alpha}N_g(k)N_s(k) \\
  & - & \alpha (M_0-N_g(k)N_s(k)  +  \alpha_w N_s(k-T_a) \\
  & + & \alpha_r N_s(k-T_{ia}) + \tilde{\alpha}N_g(k-T_{ca})N_s(k-T_{ca})\\
  & + & \tilde{\alpha}N_g(k-T_{cda})N_s(k-T_{cda})\\
  & + & \alpha(M_0-N_g(k-T_{cda}))N_s(k-T_{cga})\Gamma(k,T_{ga}).
\end{eqnarray*}
Here, $\alpha_w$ and $\alpha_r$ are the rates at which robots
encounter a wall or another robot. The current time step is $k$,
$k=0, 1, 2, \ldots$, and $T_{xyz}=T_x+T_y+T_z$ are the number of
time steps required to complete actions such as avoidance ($T_a$),
success dance ($T_d$) or stick centering ($T_c$).
$\Gamma(k,T_{ga})=\Pi_{j=k-T_{ga}}^{k}{[1-\tilde{\alpha}N_s(j)}\Theta(k-T_{ga})$
is the discrete time analog of Eq.~\ref{fc2}.

\begin{figure}[tp]
\epsfxsize = 3.2in \center{\epsffile{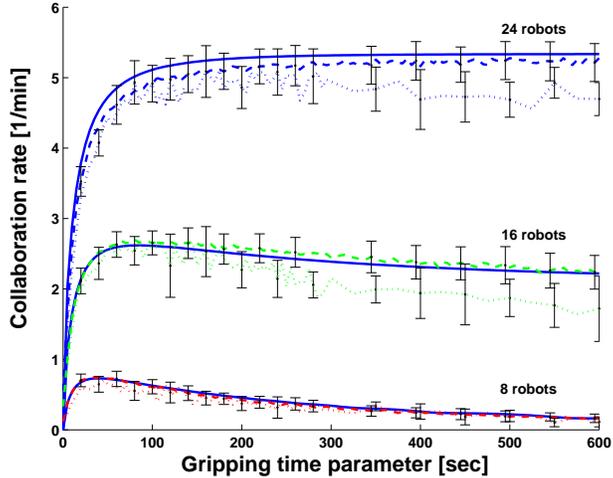}}
\vskip 0.1in \caption{ Collaboration rate as a function of the
gripping time parameter for several robot group sizes. Results are
for embodied simulations (dotted lines), the microscopic model
(dashed lines), and the macroscopic model (solid lines) in the 16
stick 80 cm radius arena. (Figure courtesy of A.~Martinoli) }
\label{fig:collaboration-difference}
\end{figure}

Figure \ref{fig:collaboration-difference} shows collaboration rate
as a function of the gripping time parameter for different robot
group sizes. Collaboration rate is computed from the number of
extracted sticks per unit time, as before:
$C(k)=\tilde{\alpha}N_s(k-T_{cd})N_g(k-T_{cd})$. We can see that
for groups as small as 8 robots, the results of the macroscopic
model quantitatively agree with embodied and microscopic
simulations.

\comment{
\section{Related Work}
\label{sec:related}

With the exceptions noted below, there has been very little prior
work on mathematical analysis of multi-agent systems. The closest
in spirit to our paper is the work by Huberman, Hogg and coworkers
on computational ecologies~\cite{HubermanHogg88,Kephart90}. These
authors mathematically studied collective behavior in a system of
agents, each choosing between two alternative strategies. They
derived a rate equation for the average number of agents using
each strategy from the underlying probability distributions. Our
approach is consistent with theirs --- in fact, we can easily
write down the same rate equations from the macroscopic state
diagram of the system, without having to derive them from the
underlying probability distributions. Computational ecologies can,
therefore, be considered an application of the methodology
described in this paper. Yet another application of the approach
presented here is the author's work on coalition formation in
electronic marketplaces \cite{Lerman00a}.

In the robotics domain, Sugawara and
coworkers~\cite{Sugawara97,SugSanYosAbe98} developed simple
state-based analytical models of cooperative foraging in groups of
communicating  and non-co\-mmu\-ni\-cating robots and studied them
quantitatively. Although these models are similar to ours, they
are overly simplified and fail to take crucial interactions among
robots into account. In separate papers, we have analyzed
collaborative~\cite{Lerman01a} and foraging~\cite{Lerman02a}
behavior in groups of robots. The focus of that work is on
realistic models and the comparison of the models' predictions to
experimental and simulations results. For example, in
~\cite{Lerman01a}, we considered the same model of collaborative
stick-pulling presented here, but studied it under the same
conditions as the experiments. In \cite{Lerman02a}, we found that
we had to include avoiding-while-searching and wall-avoiding
states in the model in order to obtain good quantitative agreement
between the model and results of sensor-based simulations. The
focus of this paper, on the other hand, is to show that there is a
principled way to construct a macroscopic model of collective
dynamics of a MAS, and, more importantly, a practical ``recipe''
for creating such a model from the details of the microscopic
controller.

}

\section{Limitations of this approach}

The rate equation approach we presented is valid for Markov and
semi-Markov systems, in which the agent's future state depends
only on its present state and, for semi-Markov systems, on how
much time it has spent in that state, and not on any of the past
states. While many systems, including reactive and behavior-based
robotics and some software agent systems, clearly obey the Markov
property, other systems composed of agents with memory, learning
or deliberative capabilities do not, and therefore, cannot be
described by the simple models presented here. However, the rate
equations are useful for studying systems of simple agents.
Moreover, with some additional complexity, this approach can be
extended to cover a broader range of agent behaviors. For
instance, in simple learning scenarios, agents could adjust their
behaviors based on the number of times they enter certain states.
For instance, to estimate the density of objects or other robots
in their environment based on the number of encounters. Thus the
transition rates would depend on the agent histories. Although the
rate equations wouldn't model the learning process per se, it
could include the effect on system behavior of improved
discrimination ability. Such a model could indicate the likely
tradeoff between exploring the environment to improve parameter
estimates and exploit those estimates for the task at hand.

Another potential limitation of the approach is that it is best as
a description of large systems. The benefit of working with large
systems is that, usually, one may safely ignore fluctuations.
Although it is possible to use the Master Equation to derive the
equation for the fluctuations about the average quantities, in
practice it is too algebraically messy. Fortunately, there exists
empirical evidence \cite{Pacala,MartinoliEaston02} that
approximate average models can provide a good quantitative
description of systems as small as a dozen agents. Large
collections of very simple agents are also seen to be reasonably
described by relations among a few aggregate
variables~\cite{bohringer97}.

Developing a suitable stochastic model for a multi-agent system
requires an understanding of the environment and agent actions
sufficient to determine the appropriate state description and
resulting transition rates. More generally, this issue can be
viewed as identifying an appropriate statistical ensemble, i.e.,
set of states with associated probabilities for their
occurrence~\cite{hogg87PhysRep}. Even oversimplified models can be
useful in giving qualitative understanding of the likely design
tradeoffs prior to a more detailed evaluation via simulation or
experiments. Nevertheless, it can be difficult to determine the
transition rates between states, especially if they involve
correlated activities among several agents. Along these lines is
the question of to what extent a detailed model of the agent
behaviors must be included. Complex agent programs could give rise
to complex transition rates that make the analysis described here
impractical. However, even in those cases, there may be some
aggregate behaviors that can be approached by suitable
coarse-graining. An extreme example is economics, which predicts
various aggregate human behaviors without requiring detailed
psychological models.

Beyond these general limitations of the stochastic approach
described here, we introduced several simplifying assumptions.
While these are not strictly necessary for the validity of the
overall approach, they are important for producing analytically
simple models. In particular we suppose the interesting system
behavior is governed by averages and the transition rates are
spatially uniform so it is not necessary to include position as
part of the state. We also extensively use the mean-field
approximation. In cases where this is not sufficient, more
accurate approximations of our statistical models are
possible~\cite{opper01}, but are more difficult to evaluate.

When evaluating the suitability of stochastic models, in addition
to these technical limitations, it is important to note the kinds
of results the models can deliver. In particular, they address
properties of the distribution of outcomes, \eg, average and
variance, over a set of repeated experiments. This is often
appropriate for evaluating how a multi-agent system will likely
perform for a class of problems. However, if one is interested in
worst-case bounds, the behavior in exceptional situations or
extreme values of the distribution (\eg, time required to find the
first or last object in a search scenario), these stochastic
techniques are unlikely to provide much insight unless the
distributions happen to be limited to a narrow range. Thus the
usefulness of these models depends not only on the complexity of
the agent behaviors but also on the nature of the collective
properties of interest.

\section{Conclusion}
\label{sec-Conclusion} We have presented an overview of
mathematical approaches for modeling and analysis of multi-agent
systems, focusing on macroscopic analytic models. Moreover, we
have described a general methodology for mathematical analysis of
such systems. Our analysis applies to a class of systems known as
stochastic Markov systems. They are stochastic, because the
behavior of each agent is inherently probabilistic in nature and
unpredictable, and they are Markovian, because the state of an
agent at a future time depends only on the present state (and
perhaps on much time the agent has spent in this state) and not on
any past state. Though each agent is unpredictable, the
probabilistic description of the collective behavior is
surprisingly simple. Our mathematical approach is based on the
stochastic Master Equation, and on the Rate Equations derived from
it, that describe how the average macroscopic, or collective,
system properties change in time. In order to create a
mathematical model, one needs to account for every relevant state
of the multi-agent system as well as for transitions between
states. For each state there is a dynamical variable in the
mathematical model and a rate equation that describes how the
variable changes in time.

We illustrated the mathematical formalism by applying it to study
collective behavior of robotic systems. Even the simplest type of
dimensional analysis of the equations yields important insights
into the system, such as what are the important parameters that
determine the behavior of the system.

In the applications, we focused on the simplest mathematical
models ({\em i.e.,} those using the smallest possible number of
states) that capture the salient features of each system. These
simple models provide a good description of the $qualitative$
behavior of the system, but in order to also obtain good
$quantitative$ agreement with experiment or simulations, it is
often necessary to include more states in the model. The approach
presented here can be easily extended to describe heterogeneous
agent systems. As a simple example, consider two, possibly
interacting, populations of foraging robots, each described by
different physical parameters. The model of a heterogeneous
foraging system will consist of two sets of coupled differential
equations, one for each sub-population, possibly with couplings
between them to represent interactions between the two
populations. It is trivial to extend the analysis to more than two
populations.

The models also provide design guidelines: choosing the agent
behaviors to closely match the simplifying assumptions of this
approach allows evaluating the resulting collective behavior via
the rate equations.

\section*{Acknowledgements}
The authors thank Dani Goldberg, Bernardo Huberman, Jeff Kephart,
Alcherio Martinoli, Maja Matari{\'c}, Richard Ross and Onn Shehory
for useful discussions.

%

\begin{thebibliography}{10}

\bibitem{abelson99}
Harold Abelson et~al.
\newblock Amorphous computing.
\newblock {\em Communications of the ACM}, 43:74--82, May 2000.

\bibitem{Agassounon02}
William Agassounon and Alcherio Martinoli.
\newblock A macroscopic model of an aggregation experiment using embodied
  agents in groups of time-varying sizes.
\newblock In {\em Proc. of the IEEE Conf. on System, man and Cybernetics
  SMC-02, October 2002, Hammamet, Tunisia. To appear.} 2002.

\bibitem{Arkin}
Ronald~C. Arkin.
\newblock {\em Behavior-Based Robotics}.
\newblock The MIT Press, Cambridge, MA, 1999.

\bibitem{BarabasiStanley95}
A.-L. Barabasi and H.E. Stanley.
\newblock {\em Fractal Concepts in Surface Growth}.
\newblock Cambridge University Press, 1995.

\bibitem{Beckers94}
R.~Beckers, O.~E. Holland, and J.~L. Deneubourg.
\newblock From local actions to global tasks: Stigmergy and collective
  robotics.
\newblock In Rodney~A. Brooks and Pattie Maes, editors, {\em Proceedings of the
  4th International Workshop on the Synthesis and Simulation of Living Systems
  \(Artificial Life {IV}\)}, pages 181--189, Cambridge, MA, USA, July 1994. MIT
  Press.

\bibitem{bender78}
Carl~M. Bender and Steven~A. Orszag.
\newblock {\em Advanced Mathematical Methods for Scientists and Engineers}.
\newblock McGraw Hill, NY, 1978.

\bibitem{Beni88}
G.~Beni.
\newblock The concept of cellular robotics.
\newblock In {\em Proc. of the 1988 IEEE Int. Symp. on Intelligent Control},
  pages 57--62, Los Alamitos, CA, 1988. IEEE Computer Society Press.

\bibitem{berlin97}
Andrew~A. Berlin and Kaigham~J. Gabriel.
\newblock Distributed {MEMS}: New challenges for computation.
\newblock {\em Computational Science and Engineering}, 4(1):12--16,
  January-March 1997.

\bibitem{bohringer97}
Karl~F. Bohringer et~al.
\newblock Computational methods for design and control of {MEMS}
  micromanipulator arrays.
\newblock {\em Computational Science and Engineering}, 4(1):17--29,
  January-March 1997.

\bibitem{bojinov02}
Hristo Bojinov, Arancha Casal, and Tad Hogg.
\newblock Multiagent control of modular self-reconfigurable robots.
\newblock {\em Artificial Intelligence}, 142:99--120, 2002.
\newblock Preprint available at {Los Alamos} archive cs.RO/0006030.

\bibitem{Bonabeau97}
Eric Bonabeau.
\newblock From classical models of morphogenesis to agent-based models of
  pattern formation.
\newblock {\em Artifical Life}, 3:191--211, 1997.

\bibitem{BDT99}
Eric Bonabeau, Marco Dorigo, and Guy Theraulaz.
\newblock {\em Swarm Intelligence: From Natural to Artificial Systems}.
\newblock Oxford University Press, New York, 1999.

\bibitem{CSS00}
D.~Chowdhury, L.~Santen, and A.~Schadschneider.
\newblock Statistical physics of vehicular traffic and some related systems.
\newblock {\em Physics Reports}, 329:199, 2000.

\bibitem{clearwater96}
Scott~H. Clearwater, editor.
\newblock {\em Market-Based Control: A Paradigm for Distributed Resource
  Allocation}.
\newblock World Scientific, Singapore, 1996.

\bibitem{courtois85}
P.~J. Courtois.
\newblock On time and space decomposition of complex structures.
\newblock {\em Communications of the ACM}, 28(6):590--603, June 1985.

\bibitem{Cushing}
J.~M. Cushing.
\newblock {\em An Introduction to Structured Population Dynamics}, volume~71 of
  {\em CBMS-NSF Regional Conference Series in Applied Mathematics}.
\newblock Society for Applied and Industrial Mathematics, Philadelphia, 1998.

\bibitem{Deneubourg92}
Jean-Luis Deneubourg, Guy Theraulaz, and R.~Beckers.
\newblock Swarm-made architectures. in: Toward a practice of autonomous
  systems.
\newblock In F.J. Varela and P.~Bourgine, editors, {\em Proceedings of The
  First European Conference on Artificial Life}, pages 123--133, Cambridge, MA,
  1992. MIT Press/Bradford Books.

\bibitem{epidemiology}
O.~Diekmann and J.~A.~P. Heesterbeek.
\newblock {\em Mathematical Epidemiology of Infectious Diseases : Model
  Building, Analysis and Interpretation}.
\newblock John Wiley \& Sons, New York, mathematical and computational biology
  edition, April 2000.

\bibitem{RobDuncan}
R.~Dun\-can et~al.
\newblock Statistical paradigms for robotic swarm modeling.
\newblock http://www.mhpcc.edu/research/ab98/98ab41.html, 1998.

\bibitem{FonMat96}
Miguel~Schneider Fontan and Maja~J Matari{\'c}.
\newblock A study of territoriality: The role of critical mass in adaptive task
  division.
\newblock In P.~Maes, M.~J. Matari{\'c}, J.~A. Meyer, J.~Pollack, and
  S.~Wilson, editors, {\em From Animals to Animats 4: Proceedings of the 4th
  International Conference on Simulation of Adaptive Behavior}, pages 553--561,
  Cambridge, MA, 1996. MIT Press.

\bibitem{freitas99}
Robert~A. {Freitas Jr.}
\newblock {\em Nanomedicine}, volume~1.
\newblock Landes Bioscience, Georgetown, TX, 1999.
\newblock Available at www.nanomedicine.com.

\bibitem{Gardiner}
C.~W. Garnier.
\newblock {\em Handbook of Stochastic Methods}.
\newblock Springer, New York, NY, 1983.

\bibitem{StagePlayer}
Brian~P. Gerkey, Richard~T. Vaughan, Kasper Støy, Andrew Howard,
Gaurav~S.
  Sukhatme, and Maja~J Matari{\'c}.
\newblock Most valuable player: A robot device server for distributed control.
\newblock In {\em Proc. of the {IEEE/RSJ} International Conference on
  Intelligent Robots and Systems (IROS 2001), Wailea, Hawaii, October 29 -
  November 3, 2001}. 2001.
\newblock http://www-robotics.usc.edu/player/.

\bibitem{GolMat00}
Dani Goldberg and Maja~J Matari{\'c}.
\newblock Robust behavior-based control for distributed multi-robot collection
  tasks.
\newblock Technical Report IRIS-00-387, USC Institute for Robotics and
  Intelligent Systems, 2000.

\bibitem{ecology}
W.~S.~C. Gurney and R.~M. Nisbet.
\newblock {\em Ecological Dynamics}.
\newblock Oxford University Press, New York, NY, 1998.

\bibitem{Haberman}
Richard Haberman.
\newblock {\em Mathematical Models: Mechanical Vibrations, Population Dynamics,
  and Traffic Flow}.
\newblock Society of Industrial and Applied Mathematics ({SIAM}), Philadelphia,
  PA, 1998.

\bibitem{Helbing}
Dirk Helbing.
\newblock {\em Quantitative Sociodynamics: Stochastic Methods and Models of
  Social Interaction Processes}, volume~31 of {\em THEORY AND DECISION LIBRARY
  B: Mathematical and Statistical Methods}.
\newblock Kluwer Academic, Dordrecht, 1995.

\bibitem{HelbingSchweitzer}
Dirk Helbing, Frank Schweitzer, Joachim Keltsch, and Peter Molnar.
\newblock Active walker model for the formation of human and animal trail
  systems.
\newblock {\em Physical Review}, E 56(3):2527--2539, 1997.

\bibitem{hirshleifer78}
J.~Hirshleifer.
\newblock Competition, cooperation, and conflict in economics and biology.
\newblock {\em The American Economic Review}, 68(2):238--243, May 1978.

\bibitem{hogg87PhysRep}
T.~Hogg and B.~A. Huberman.
\newblock Artificial intelligence and large scale computation: A physics
  perspective.
\newblock {\em Physics Reports}, 156:227--310, 1987.

\bibitem{hogg96d}
Tad Hogg, Bernardo~A. Huberman, and Colin~P. Williams, editors.
\newblock {\em Frontiers in Problem Solving: Phase Transitions and Complexity},
  volume~81, Amsterdam, 1996. Elsevier.
\newblock Special issue of {\it Artificial Intelligence}.

\bibitem{HolMel00}
Owen Holland and Chris Melhuish.
\newblock Stigmergy, self-organization and sorting in collective robotics.
\newblock {\em Artificial Life}, 5(2), 2000.

\bibitem{huberman93a}
Bernardo~A. Huberman and Natalie~S. Glance.
\newblock Evolutionary games and computer simulations.
\newblock {\em Proceedings of the National Academy of Science USA},
  90:7716--7718, August 1993.

\bibitem{HubermanHogg88}
Bernardo~A. Huberman and Tad Hogg.
\newblock {The behavior of computational ecologies}.
\newblock In B.~A. Huberman, editor, {\em {The Ecology of Computation}}, pages
  77--115, Amsterdam, 1988. Elsevier (North-Holland).

\bibitem{IMB2001}
A.~J. Ijspeert, A.~Martinoli, A.~Billard, and L.~M. Gambardella.
\newblock Collaboration through the exploitation of local interactions in
  autonomous collective robotics: The stick pulling experiment.
\newblock {\em Autonomous Robots}, 11(2):149--171, 2001.

\bibitem{Kazadi02}
Sanza Kazadi, A.~Abdul-Khaliq, and Ron Goodman.
\newblock On the convergence of puck clustering systems.
\newblock {\em Robotics and Autonomous Systems}, 38(2):93--117, 2002.

\bibitem{Kephart90}
Jeffrey~O. Kephart, Tad Hogg, and Bernardo~A. Huberman.
\newblock Collective behavior of predictive agents.
\newblock {\em Physica}, D 42:48--65, 1990.

\bibitem{KitTamStoVel98}
H.~Kitano, M.~Tambe, P.~Stone, and M.~Veloso.
\newblock The {RoboCup} synthetic agent challenge 97.
\newblock {\em Lecture Notes in Computer Science}, 1395:62--??, 1998.

\bibitem{KubBon00}
C.~R. Kube and E.~Bonabeau.
\newblock Cooperative transport by ants and robots.
\newblock {\em Robotics and Autonomous Systems}, 30(1--2):85--101, 2000.

\bibitem{Lerman02a}
Kristina Lerman and Aram Galstyan.
\newblock Mathematical model of foraging in a group of robots: Effect of
  interference.
\newblock {\em Autonomous Robots}, 13(2):127--141, 2002.

\bibitem{Lerman01a}
Kristina Lerman, Aram Galstyan, Alcherio Martinoli, and Auke
Ijspeert.
\newblock A macroscopic analytical model of collaboration in distributed
  robotic systems.
\newblock {\em Artificial Life Journal}, 7(4):375--393, 2001.

\bibitem{Lerman00a}
Kristina Lerman and Onn Shehory.
\newblock {Coalition Formation for Large-Scale Electronic Markets}.
\newblock In {\em Proceedings of the International Conference on Multi-Agent
  Systems ({ICMAS}'2000), Boston, MA, 2000.}, pages 167--174, 2000.

\bibitem{Mahnke99}
R.~Mahnke and J.~Kaupu\^zs.
\newblock Stochastic theory of freeway traffic.
\newblock {\em Physical Review}, E59(1):117--125, 1999.

\bibitem{Mahnke97}
R.~Mahnke and N.~Pieret.
\newblock Stochastic master-equation approach to aggregation in freeway
  traffic.
\newblock {\em Physical Review}, E56(3):2666--2671, 1997.

\bibitem{Martinoli99}
A.~Martinoli.
\newblock {\em Swarm Intelligence in Autonomous Collective Robotics: From Tools
  to the Analysis and Synthesis of Distributed Control Strategies}.
\newblock PhD thesis, PhD Thesis No 2069, EPFL, 1999.

\bibitem{MarIjsGam99}
A.~Martinoli, A.~J. Ijspeert, and L.~M. Gambardella.
\newblock A probabilistic model for understanding and comparing collective
  aggregation mechanisms.
\newblock In Dario Floreano, Jean-Daniel Nicoud, and Francesco Mondada,
  editors, {\em Proceedings of the 5th European Conference on Advances in
  Artificial Life ({ECAL}-99)}, volume 1674 of {\em LNAI}, pages 575--584,
  Berlin, September~13--17 1999. Springer.

\bibitem{MarMon95}
A.~Martinoli and F.~Mondada.
\newblock Collective and cooperative group behaviors: Biologically inspired
  experiments in robotics.
\newblock In O.~Khatib and J.~K. Salisbur, editors, {\em Proc. of the Fourth
  Int. Symp. on Experimental Robotics ISER-95}. Springer Verlag, June-July
  1995.

\bibitem{MartinoliISER02}
Alcherio Martinoli and Kjerstin Easton.
\newblock Modeling swarm robotic systems.
\newblock In B.~Siciliano and P.~Dario, editors, {\em Proc. of the Eight Int.
  Symp. on Experimental Robotics ISER-02, Sant'Angelo d'Ischia, Italy},
  Springer Tracts in Advanced Robotics 5, pages 297--306, New York, NY, july
  2003. Springer Verlag.

\bibitem{MartinoliEaston02}
Alcherio Martinoli and Kjerstin Easton.
\newblock Optimization of swarm robotic systems via macroscopic models.
\newblock In A.~C. Schultz, L.~E. Parker, and F.~E. Schneider, editors, {\em
  Proc. of the Second Int. Workshop on Multi-Robots Systems, March, 2003,
  Washington, DC}, pages 181--192. Kluwer Academic Publishers, 2003.

\bibitem{Mataric92}
M.~Matari{\'c}.
\newblock Minimizing complexity in controlling a mobile robot population.
\newblock In {\em Proceedings of the 1992 IEEE International Conference on Robo
  tics and Automation}, pages 830--835, Nice, France, 1992.

\bibitem{MatNilSim95}
M.~J. Matari{\'c}, M.~Nilsson, and K.~Simsarian.
\newblock Cooperative multi-robot box pushing.
\newblock In {\em Proceedings of the 1995 IEEE/RSJ International Conference on
  Intelligent Robots}, 1995.

\bibitem{Melhuish98}
C.~Melhuish, O.~Holland, and S.~Hoddell.
\newblock Collective sorting and segregation in robots with minimal sensing.
\newblock In R.~Pfeifer, B.~Blumberg, J.-A. Meyer, and S.W. Wilson, editors,
  {\em {From Animals to Animats, Proceedings of the Fifth International
  Conference of The Society for Adaptive Behavior (SAB98)}}, pages 465--470.
  MIT Press, 1998.

\bibitem{Michel98}
O.~Michel.
\newblock Webots: Symbiosis between virtual and real mobile robots.
\newblock In J.-C. Heudin, editor, {\em {Proc. of the First Int. Conf. on
  Virtual Worlds, Paris, France,}}, pages 254--263. Springer Verlag, 1998.
\newblock See also http://www.cyberbotics.com/webots/.

\bibitem{ArkBalNit}
E.~Nitz, R.~C. Arkin, and T.~Balch.
\newblock Communication of behavioral state in multi-agent retrieval tasks.
\newblock In Lisa {Werner, Robert; O'Conner}, editor, {\em Proceedings of the
  1993 {IEEE} International Conference on Robotics and Automation: Volume 3},
  pages 588--594, Atlanta, GE, May 1993. IEEE Computer Society Press.

\bibitem{opper01}
Manfred Opper and David Saad, editors.
\newblock {\em Advanced Mean Field Methods: Theory and Practice}.
\newblock MIT Press, Cambridge, MA, 2001.

\bibitem{OstSukMat01}
Esben~H. {\O}stergaard, Gaurav~S. Sukhatme, and Maja~J.
Matari{\'c}.
\newblock Emergent bucket brigading - a simple mechanism for improving
  performance in multi-robot constrained-space foraging tasks.
\newblock In {\em Proceedings of the 5th International Conference on Autonomous
  Agents ({AGENTS}-01)}, 2001.

\bibitem{Pacala}
Stephen~W. Pacala, Deborah~M. Gordon, and H.~C.~J. Godfray.
\newblock Effects of social group size on information transfer and task
  allocation.
\newblock {\em Evolutionary Ecology}, 10:127--165, 1996.

\bibitem{Parker98}
Lynne Parker.
\newblock Alliance: An architecture for fault-tolerant multi-robot cooperation.
\newblock {\em IEEE Transactions on Robotics and Automation}, 14(2):220--240,
  1998.

\bibitem{purcell77}
E.~M. Purcell.
\newblock Life at low reynolds number.
\newblock {\em American Journal of Physics}, 45:3--11, 1977.

\bibitem{Loomis99}
W.-J. Rappel, A.~Nicol, A.~Sarkissian, H.~Levine, and W.~F.
Loomis.
\newblock Self-organized vortex state in two-dimensional dictyostelium
  dynamics.
\newblock {\em Physical Review Letters}, 83:1247--1250, 1999.

\bibitem{SchoonderwoerdEtAl97}
Ruud Schoonderwoerd, Owen Holland, and Janet Bruten.
\newblock Ant-like agents for load balancing in telecommunications networks.
\newblock In W.~Lewis Johnson and Barbara Hayes-Roth, editors, {\em Proceedings
  of the 1st International Conference on Autonomous Agents}, pages 209--216,
  New York, February~5--8 1997. ACM Press.

\bibitem{Schweitzer+al97}
Frank Schweitzer, Kenneth Lao, and Fereydoon Family.
\newblock Active random walkers simulate trunk trail formation by ants.
\newblock {\em BioSystems}, 41:163--166, 1997.

\bibitem{simon96}
Herbert~A. Simon.
\newblock {\em The Sciences of the Artificial}.
\newblock MIT Press, Cambridge, MA, 3rd edition, 1996.

\bibitem{simon61}
Herbert~A. Simon and Albert Ando.
\newblock Aggregation of variables in dynamic systems.
\newblock {\em Econometrica}, 29:111--138, 1961.

\bibitem{Sugawara97}
Ken Sugawara and Masaki Sano.
\newblock Cooperative acceleration of task performance: Foraging behavior of
  interacting multi-robots system.
\newblock {\em Physica}, D100:343--354, 1997.

\bibitem{SugSanYosAbe98}
Ken Sugawara, Masaki Sano, Ikuo Yoshihara, and K.~Abe.
\newblock Cooperative behavior of interacting robots.
\newblock {\em Artificial Life and Robotics}, 2:62--67, 1998.

\bibitem{Sugawara+al99}
Ken Sugawara, Masaki Sano, Ikuo Yoshihara, K.~Abe, and
T.~Watanabe.
\newblock Foraging behavior of multi-robot system and emergence of swarm
  intelligence.
\newblock In {\em Proc. of IEEE Int. Conf. on Systems, Man, and Cybernetics
  (SMC-99)}, pages 257--262, 1999.

\bibitem{Theraulaz98}
Guy Theraulaz, Eric Bonabeau, and Jean-Luis Deneubourg.
\newblock Response threshold reinforcement and division of labour in insect
  societies.
\newblock {\em Proc. of Royal Society of London}, Serie B:327--332, 1998.

\bibitem{VauStoSukMat00b}
Richard~T. Vaughan, Kasper St{\o}y, Gaurav~S. Sukhatme, and
Maja~J.
  Matari{\'c}.
\newblock Blazing a trail: Insect-inspired resource transportation by a robot
  team.
\newblock In {\em Proceedings of the 5th International Symposium on Distributed
  Autonomous Robotic Systems (DARS), Knoxville, TN}, 2000.

\bibitem{VauStoSukMat00}
Richard~T. Vaughan, Kasper St{\o}y, Gaurav~S. Sukhatme, and
Maja~J.
  Matari{\'c}.
\newblock Whistling in the dark: cooperative trail following in uncertain
  localization space.
\newblock In Carlos Sierra, Gini Maria, and Jeffrey~S. Rosenschein, editors,
  {\em Proceedings of the 4th International Conference on Autonomous Agents
  ({AGENTS}-00)}, pages 187--194, NY, June ~3--7 2000. ACM Press.

\bibitem{Walgraef97}
Daniel Walgraef.
\newblock {\em Spatio-Temporal Pattern Formation and with Examples from Physics
  and Chemistry and and Materials Science}.
\newblock Springer, New York, NY, 1997.

\bibitem{Wolfram}
Stephen Wolfram.
\newblock {\em Cellular Automata and Complexity}.
\newblock Addison-Wesley, Reading, Mass., 1994.

\end{thebibliography}

\end{document}